\documentclass[letterpaper, 10 pt, journal, twoside]{IEEEtran}
\usepackage[subpreambles=true]{standalone}
\usepackage{import}

\usepackage{longbox}

\usepackage{amsmath} %
\usepackage{graphicx}
\usepackage{makecell}
\usepackage{amsfonts}
\usepackage{amssymb,bm}

\usepackage{xifthen}
\usepackage{pdfpages}
\usepackage{transparent}
\usepackage[]{subcaption}
\usepackage{hyperref}

\RequirePackage[loading]{tracefnt}

\usepackage[ruled,vlined,linesnumbered]{algorithm2e}
\usepackage{algpseudocode}
\usepackage{float}
\usepackage{multirow}
\usepackage{booktabs}

 \usepackage{soul}
\usepackage{stfloats}
\usepackage[belowskip=2pt,aboveskip=0pt]{caption}

\definecolor{shadecolor}{rgb}{1,.8,.3}

\newcommand \figref[1]{Fig. \ref{#1}}

\newcommand\CS{$\mathcal{C}$}
\newcommand\CSA{$\hat{\mathcal{C}}$}
\newcommand\CF{$\mathcal{C}_{free}$}
\newcommand\CFA{$\hat{\mathcal{C}}_{free}$}
\newcommand\NaturalNumbers{\mathbb{N}}

\newcommand\Real{\mathbb{R}}

\begin{document}
	\title{
Visualizing High-Dimensional Configuration Spaces: A Comprehensive Analytical Approach}

\author{Jorge Ocampo Jimenez$^{1}$ and Wael Suleiman$^{2}$, \emph{Senior Member, IEEE}%

\thanks{This work was partly supported by Consejo Nacional de Ciencia y Tecnología (CONACyT, Mexico City, Grant No. 278823) and Natural Sciences and Engineering Research Council of Canada (NSERC).}%
\thanks{$^{1}$Jorge Ocampo Jimenez and $^{2}$Wael Suleiman are with Electrical and Computer Engineering Department, Universit\'e de Sherbrooke, Quebec, Canada  (e-mail: \{Jorge.Ocampo-Jimenez; Wael.Suleiman\}@USherbrooke.ca)}%

}

\maketitle
\begin{abstract}
The representation of a Configuration Space  \CS{} plays a vital role in accelerating the finding of a collision-free path for sampling-based motion planners where the majority of computation time is spent in collision checking of states. Traditionally, planners evaluate \CS{}'s representations through limited evaluations of collision-free paths using the collision checker or by reducing the dimensionality of \CS{} for visualization. However, a collision checker may indicate high accuracy even when only a subset of the original \CS{} is represented; limiting the motion planner's ability to find paths comparable to those in the original \CS{}. Additionally, dealing with high-dimensional \CS{}s is challenging, as qualitative evaluations become increasingly difficult in dimensions higher than three, where reduced-dimensional \CS{} evaluation may decrease accuracy in cluttered environments. In this paper, we present a novel approach for visualizing representations of  high-dimensional \CS{}s of manipulator robots in a 2D format. We provide a new tool for qualitative evaluation of high-dimensional \CS{}s approximations without reducing the original dimension. This enhances our ability to compare the accuracy and coverage of two different high-dimensional \CS{}s. Leveraging the kinematic chain of manipulator robots and human color perception, we show the efficacy of our method using a 7-degree-of-freedom CS of a manipulator robot.  This visualization offers qualitative insights into the joint boundaries of the robot and the coverage of collision state combinations without reducing the dimensionality of the original data. To support our claim, we conduct a numerical evaluation of the proposed visualization.
\end{abstract}
	\IEEEpeerreviewmaketitle
\section{Introduction}\label{sec:introduction}
	\IEEEPARstart{T}{he} position and orientation of a robot can be described by a n-dimensional vector, referred to as the configuration or state. The set of all possible configurations of a robot is known as \CS{} \cite{LozanoPerez1981AutomaticPO}, where \CF{}$\subseteq$ \CS{} represents the collision-free-configurations of the robot. In a given task, the movement of a robot relies on finding a continuous curve in \CF\, from a start state to a goal state. However, in high-dimensional problems this task becomes PSPACE-hard and typically necessitates an explicit visualization of obstacles in \CS{} \cite{Karaman}. Some approaches try to simplify the problem by making assumptions about the geometry of the obstacles in \CS{} or planning in the task-space; however, it limits its utility with complex obstacles or narrow channels \cite{10.1177/0278364914528132}\cite{10.5555/1703775.1703917}. To address these challenges, sampling-based motion planners (SBMPs) have been proposed. These planners aim to avoid the explicit visualization of the entire \CS{} by randomly sampling states within it, and expect to connect the \CF{} sampled states from the start state to the goal state.

	\begin{figure}[tb]
\centering
\setlength{\fboxsep}{0pt}
	\setlength{\fboxrule}{0pt}%
	\fbox{
\begin{subfigure}[t]{.95\linewidth}
\centering
	\includegraphics[scale=0.09]{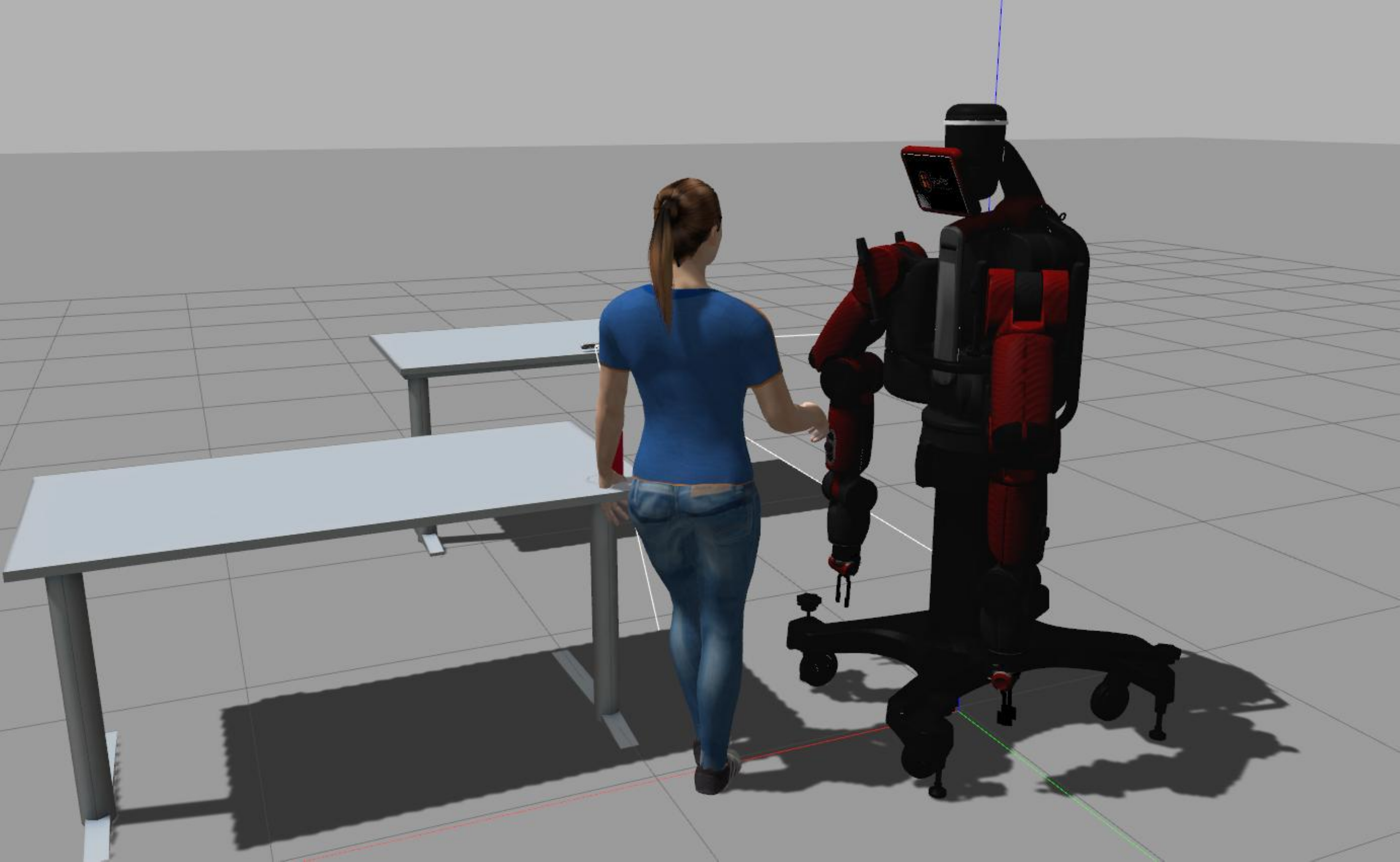}
	      \caption{WS where the human is close to the right arm of the robot. The changes in the WS are reflected in \CS{}} %
	      \label{fig:ws04}
\end{subfigure}}
\hfill
\fbox{
\begin{subfigure}[t]{.95\linewidth}
\centering
	 \includegraphics[scale=0.13]{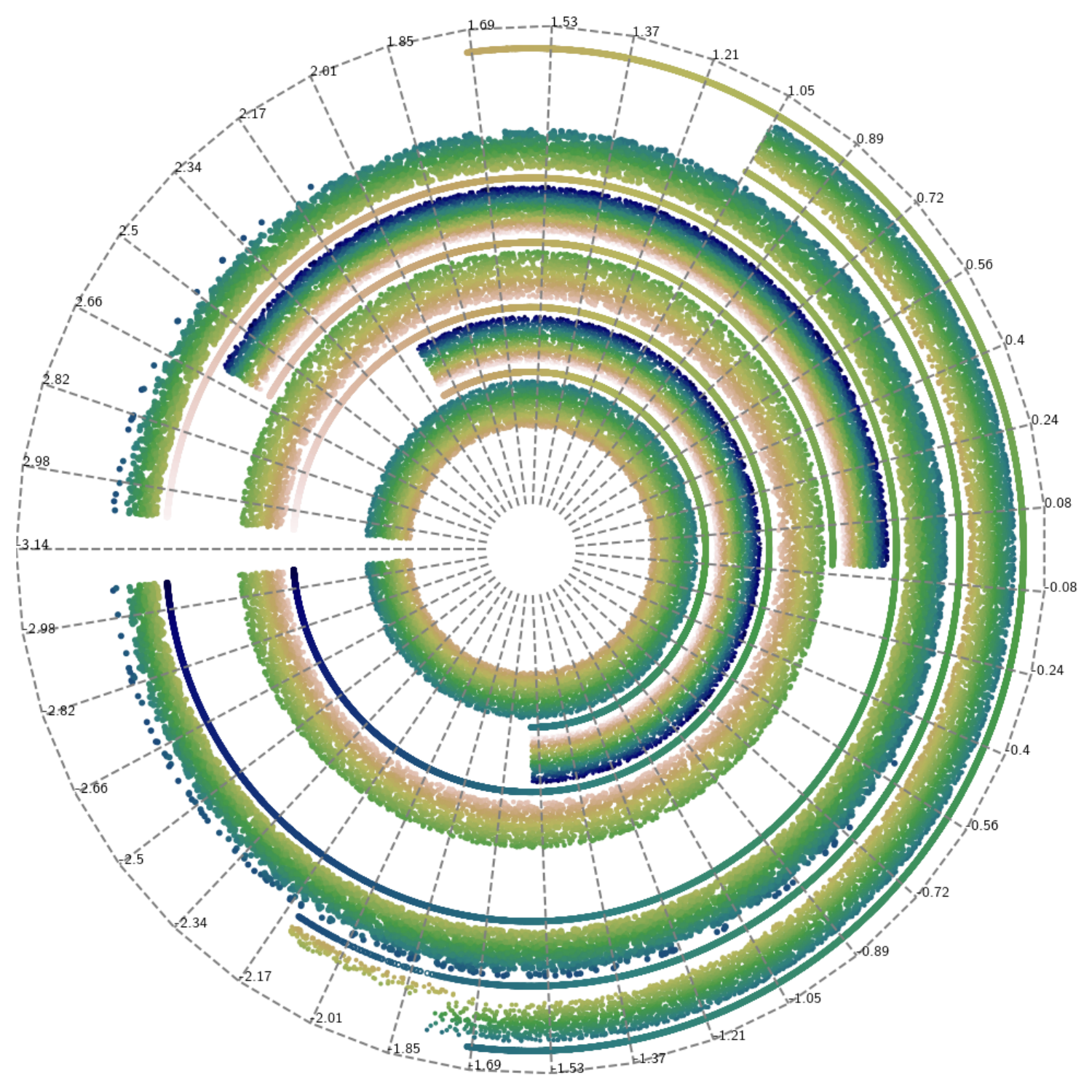}
	      \caption{Configuration Space  visualization coming from a uniform random sampler and a collision checker of the same WS}%
	      \label{fig:wholeGuides4}
\end{subfigure}}
\caption{2D visualization of a \CF{} for a 7-degree-of-freedom (DOF) robot manipulator.}
\label{fig:comparisonGS}
\end{figure}

	While SBMPs have proven highly effective in solving a wide range of problems, they pose a challenge when time constraints are imposed, particularly in real-world scenarios where the distribution of points in \CS{} differs significantly from the distribution used by the random sampler in SBMPs.

Recent years have witnessed significant advancements in manifold learning using machine learning models. SBMPs have been integrated with approximations of \CS{} that we define as \CSA{}; that account for obstacles present in the working space (WS). A learned \CS{} can serve as a biased random sampler by utilizing the states sampled from this \CFA{} as the sampling distribution of an SBMP, even without reducing the dimensionality of \CS{}. This approach has been applied to various domains, including humanoid-robot \CS{}s \cite{9385935}. Studies have demonstrated that learning the regions of interest enables finding \CF{}-paths with data-driven constraints \cite{EDI-INF-RR-1340}, resulting in computational savings by conditioning the \CSA{} based on the WS \cite{10.1109/ICRA.2018.8460730,8594028,LUO2023104545,1394271ba153437faec42ec27ffa0437,9154607,DBLP:journals/corr/abs-2303-05653}.
\begin{figure*}
	\centering
	\setlength{\fboxsep}{0pt}
	\setlength{\fboxrule}{0pt}%
	\fbox{
	\begin{subfigure}[t]{0.17\linewidth}
	    \includegraphics[scale=0.1]{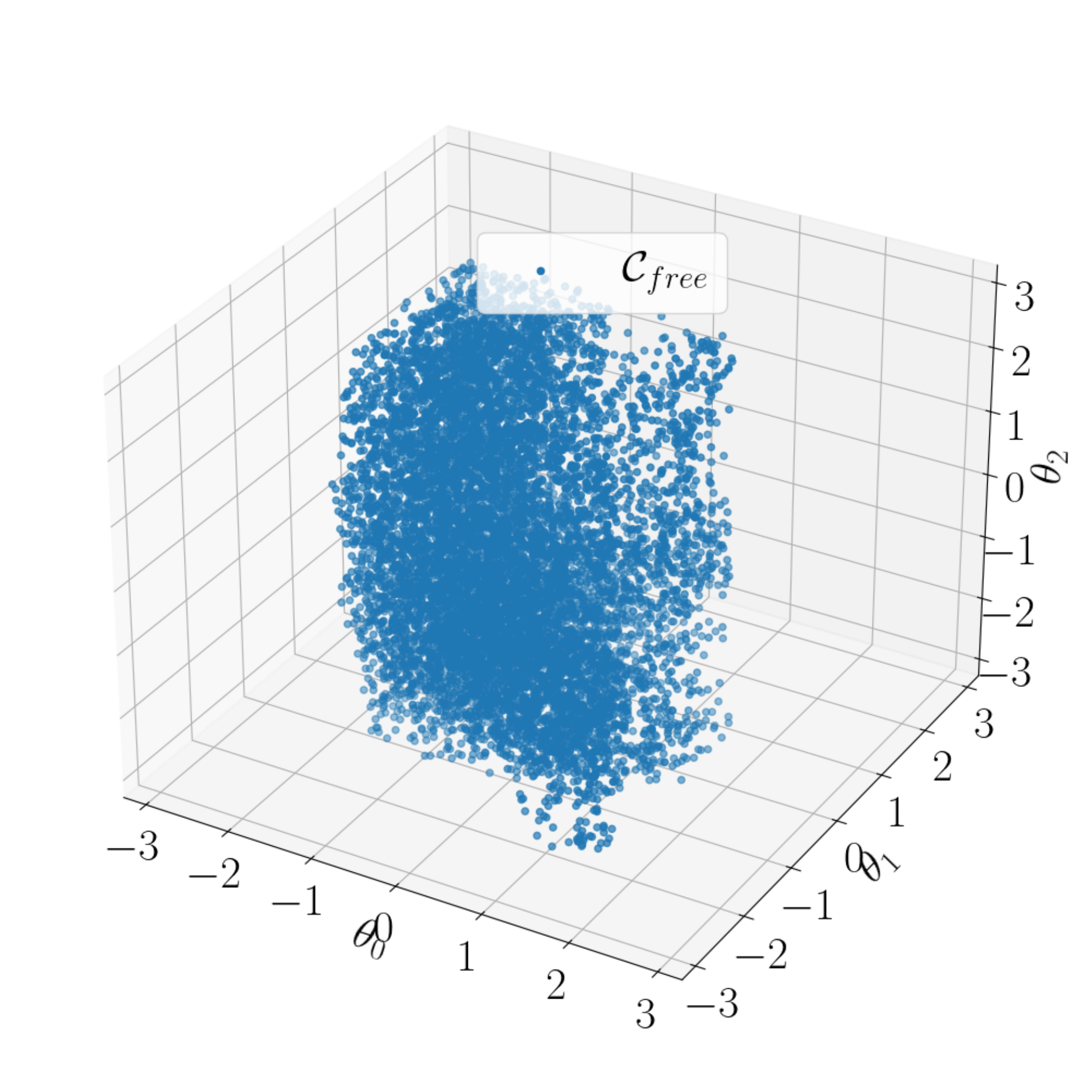}
	    \caption{3D projection.}
	     \label{fig:3DCS}
	\end{subfigure}\hspace{0\unitlength}
	}
	\fbox{
		\begin{subfigure}[t]{0.35\linewidth}
			\centering
	    \includegraphics[scale=0.10]{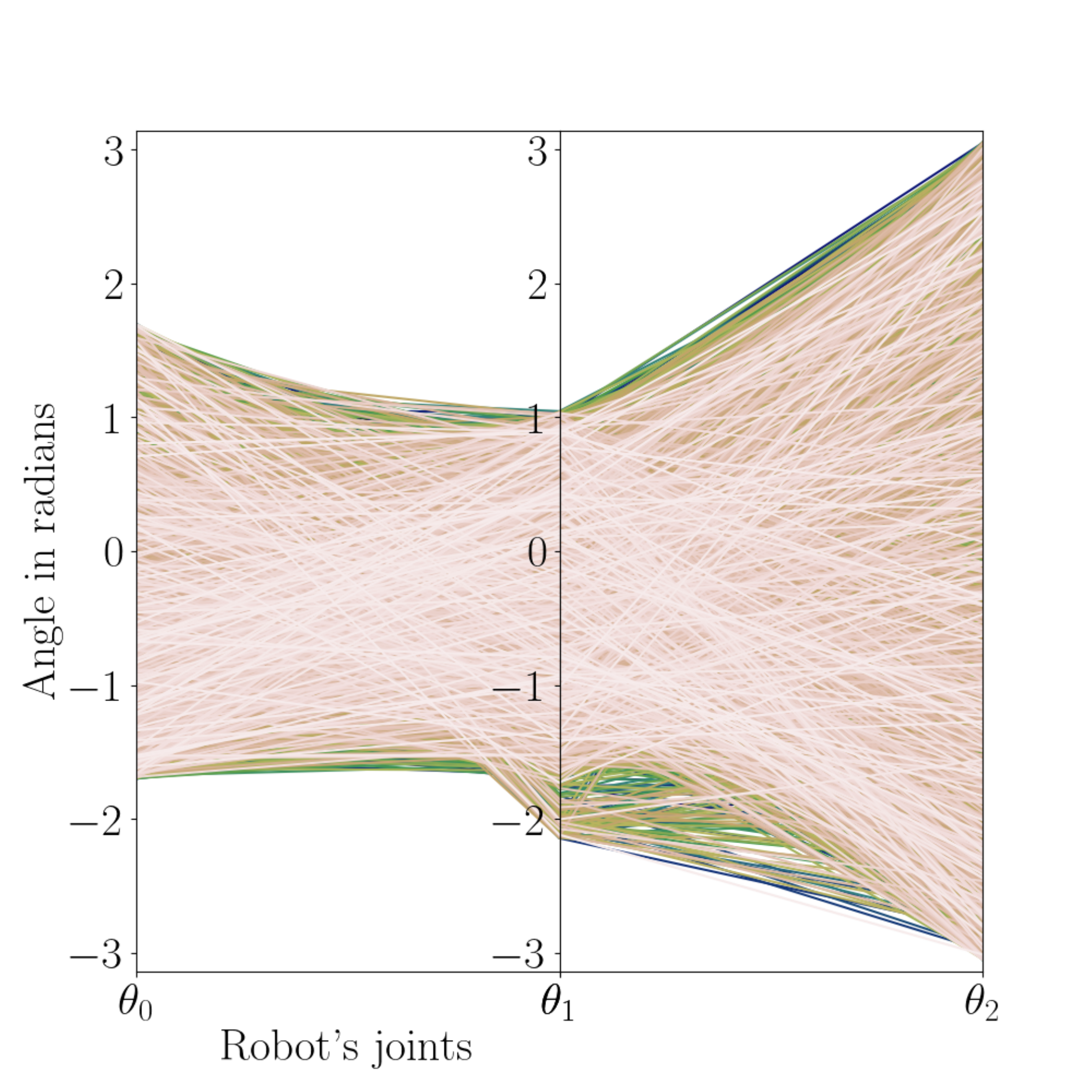}
	    \caption{Parallel coordinates.}
	     \label{fig:parallel3D}

	\end{subfigure}}
	\quad
	\begin{subfigure}[t]{0.21\linewidth}
		\centering
		\includegraphics[scale=0.32]{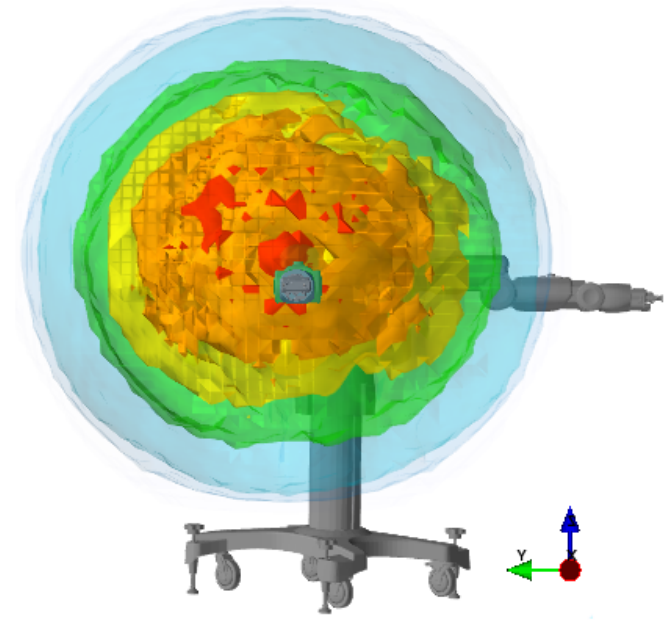}
		\caption{Projection of \CS{} in the WS.}
		\label{fig:projectionws}
	\end{subfigure}
			\raisebox{0.6\height}{
	\begin{subfigure}[t]{0.20\linewidth}
				\centering
	    \includegraphics[scale=0.06]{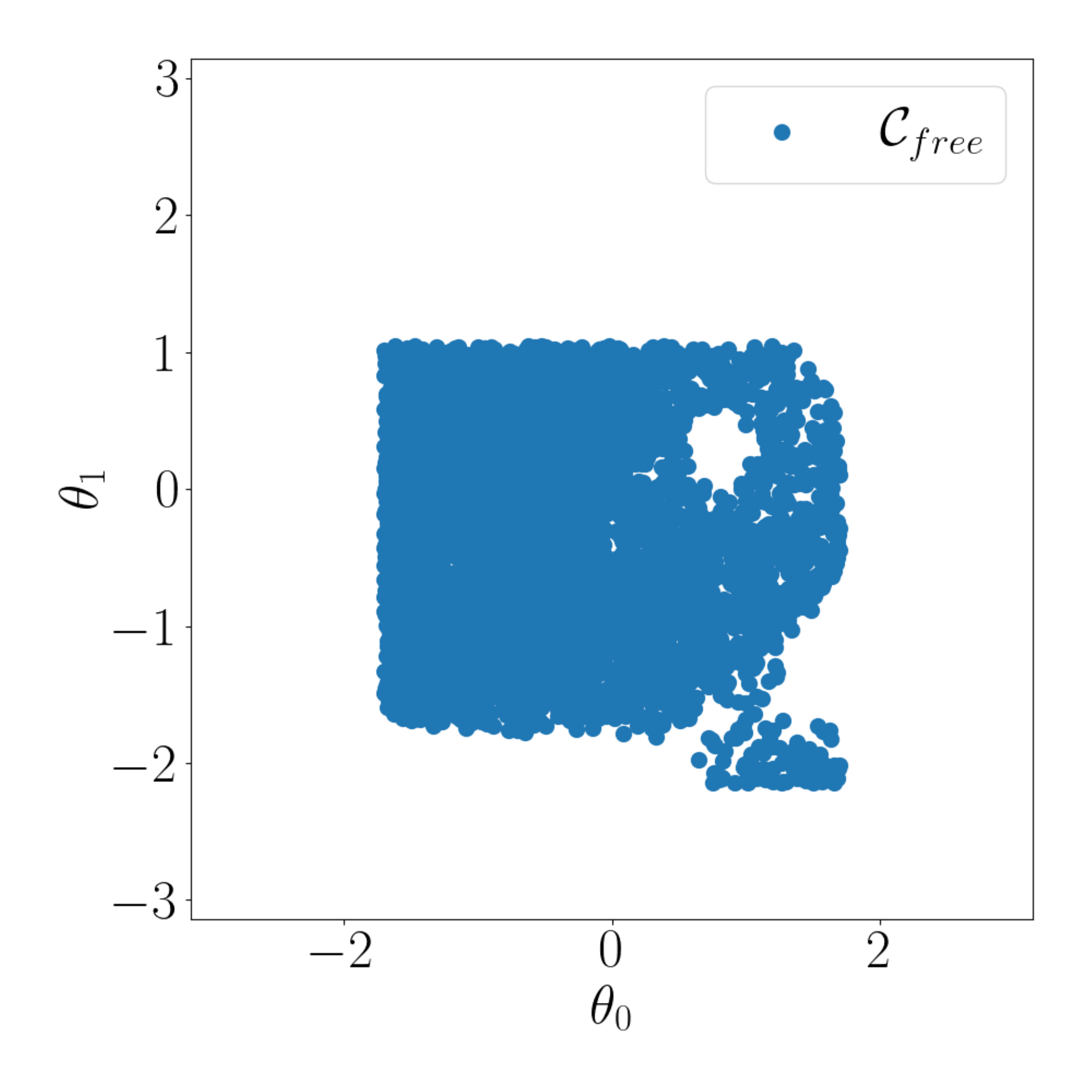}\\
		\includegraphics[scale=0.06]{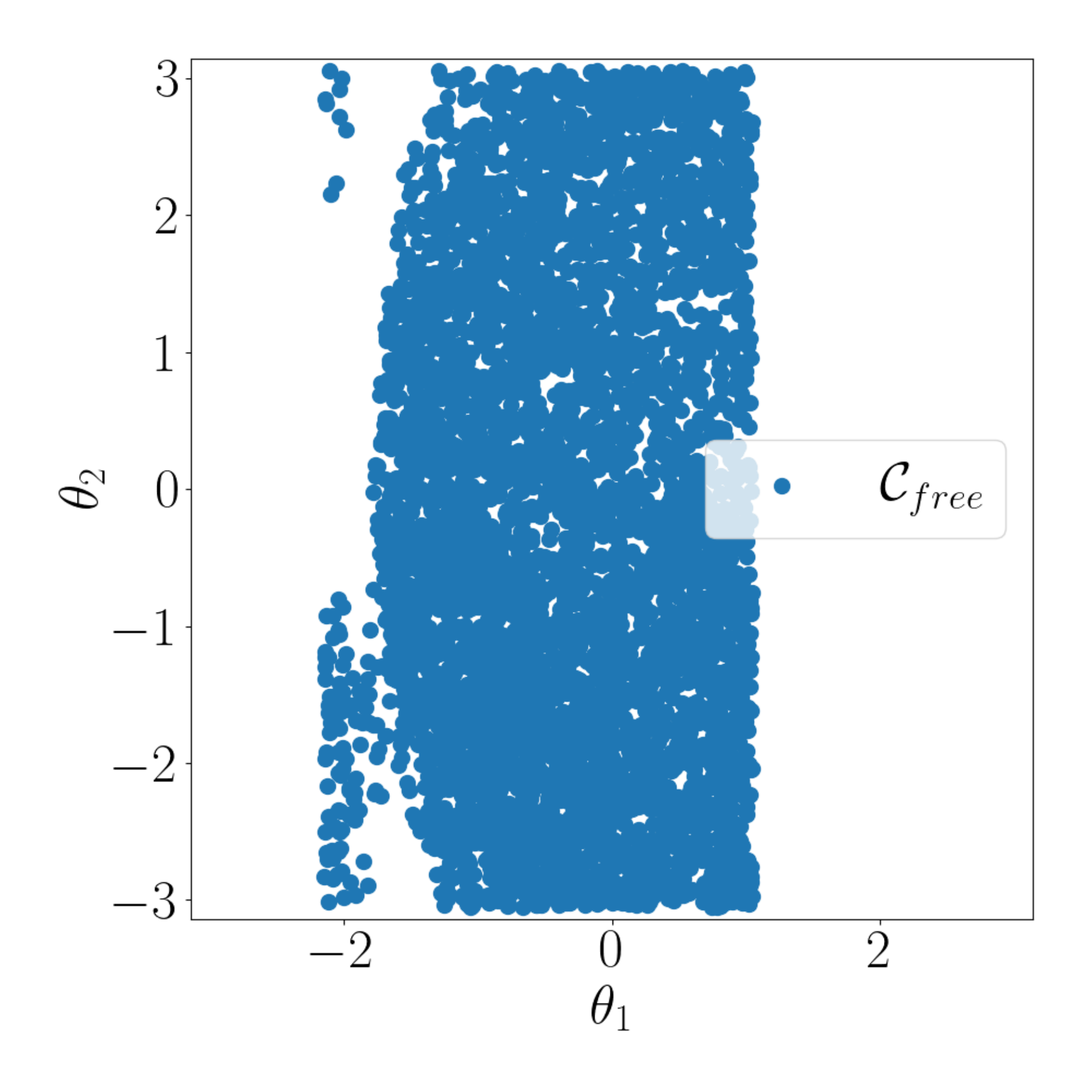}
	    \caption{2D pair projections.}
\label{fig:projection2d}
	\end{subfigure}}
\caption{Classical representations of \CS{} from the first 3 joints of a 7-DOF Baxter robot. \figref{fig:3DCS} shows 5000 \CF{}-states sampled; with joints $\theta_i \in [-\pi,\pi],i \in \NaturalNumbers$. Using parallel coordinates in \figref{fig:parallel3D} can effectively capture the constraints within individual joints by representing each independent state as a unique colored line but fails to illustrate the complex interactions between them given the produced clutter. The projection of \CS{} in WS from \figref{fig:projectionws} is useful for one sample of \CS{}; the evaluation of the whole \CS{} is difficult given that only some boundaries can be visualized when trying to show the whole distribution, the rest is occluded by the 2D projection in the 3D WS. The 2D projections in \figref{fig:projection2d}, $\theta_0 \times \theta_1$ and $\theta_1 \times \theta_2$ of \CS{} capture the interactions between adjacent joints, such as the presence of a 3D hole by projecting it in two dimensions, but the simultaneous interactions across individual samples in all dimensions can be challenging to discern in specific states/regions.}

	\label{fig:differentProjections}
	\end{figure*}

	However, qualitatively assessing the fidelity of a representation and directing sampling toward \CFA{} becomes increasingly intricate as dimensionality increases. A poor \CSA{} for the robot would necessitate additional collision checking within the SBMP framework, as it becomes necessary to verify if a state is truly in collision or not, considering the false negatives and positives provided by the learned \CSA{}. This could potentially undermine the computational savings expected from the \CSA{} during collision checking.

Most approaches to qualitatively represent \CSA{} of robots originate from the fields of computer visualization and machine learning, given the interdisciplinary nature of SBMPs. However, due to the complexity of analyzing high-dimensional-\CSA{}, only a small sample of edge cases demonstrating successful \CF{}-paths are typically provided for evaluation. Relying solely on accuracy data from the collision checker to evaluate \CFA{} can be misleading. In certain scenarios, it is possible to achieve high accuracy in \CFA{} even when only a subset of \CF{} is represented. This, in turn, can limit the number of \CF{}-paths that can be discovered. This limitation is also evident when using projections of \CFA{} onto the WS, as only a limited and local view of the representation is provided in a 2D projection of a 3D WS, while the rest of the configurations remain occluded within the projection of the joint boundaries.

Another approach to visualizing \CS{} is to reduce its dimensionality, such as through Principal Component Analysis (PCA), Linear Discriminant Analysis (LDA), or random projections to select the two or three components of \CSA{} that encode the most important information of the model. In particular, random projections overcome the limitations of other linear reduction techniques in terms of computation time. However, dimension reduction techniques often lead to data distortion, which affects accuracy in high-dimensional systems \cite{DBLP:conf/iros/SucanK09}. This issue is typically addressed by increasing the dimensionality of the \CFA{} \cite{XanthidisJIRS2020}, thus necessitating a high-dimensional-analysis. For a more comprehensive discussion of various types of data visualization, readers can refer to \cite{6064985}.

	A specific approach tailored to the field of robotics for visualizing high-dimensional \CS{}s in robotic tasks is presented in the work of \cite{100146}, where parallel coordinates are utilized to represent the data points in high-dimensional \CS{}s. However, this visualization tends to become cluttered when the number of sampling points is substantial. An example of each visualization approach is presented in \figref{fig:differentProjections}.

\section{Contributions}\label{sec:contributions}
	Given the outlined challenges from the previous section, our objective is to evaluate the representation of \CS{} both qualitatively and quantitatively before its integration into SBMP tasks by introducing a novel visualization for high-dimensional \CS{}s. The aim is to develop a visualization that offers insights into how closely a \CSA{} distribution aligns with the real distribution.

The primary contribution of this work lies in the development of a 2D graphical visualization for discrete, high-dimensional \CS{}s. This visualization serves as a valuable tool for practitioners in robotics and machine learning by aiding them in the visual identification and analysis of representations of \CF{}-regions within \CS{} by:
\begin{itemize}

    \item Visual Identification of \CF{}-regions: We introduce a novel approach that allows practitioners to visually discern the areas within high-dimensional-manipulator robot \CS{}s that are in \CF{} with a 2D projection. This visual guidance greatly facilitates a general qualitative analysis of different \CS{}s where collision regions can be detected even in high-dimensional-scenarios. The identification becomes possible by adhering to the color constraints established by the independent joints within the kinematic sequence of the joints and the bias of the distribution of the states to a specific region. This contribution is detailed in Section \ref{sec:visualization}.

    \item Qualitative Comparison of \CS{}s: Our proposed visualization offers an intuitive means to qualitatively compare two different \CSA{}s to validate the quality of a model representation of a robot \CS{}. Even in cases where two \CS{}s exhibit a global visual similarity, our visualization empowers users to employ specific operations within the visualization to identify differences in local regions, information not provided by another visualization; this is described in the Section \ref{sec:qualitative}.

    \item Encoding of Numerical Information and Evaluation Metrics: The visualization also encodes numerical information regarding how close \CSA{} is to the original \CS{}. It can also be analyzed using metrics from the field of computer vision where the differences between two different \CS{}s or subsets can be measured. This contribution is detailed in Section \ref{sec:quantitative}.
\end{itemize}

\section{Visualization Construction}\label{sec:rec}
	Constructing visualizations of \CSA{}s could be fundamental for understanding how well a model fits the original data for SBMPs. In this section, we detail the methodology employed to construct our new visualization.

Our objective is to capture the essential characteristics of a rotative manipulator robot for SBMP tasks. This entails providing a comprehensive view of \CF{} positions of the robot's joints in response to variations in the robot's current state within the WS. Additionally, our visualization enables us to detect the constraints imposed by obstacles in the environment, thereby enhancing our understanding of \CS{}.

	The process involves establishing a ground plane that connects all projections from unidimensional $\theta_{i+1}$ to $S^1$, preserving the relationship with the ground plane $\theta_{i}$. This ground plane can be likened to a coordinate chart $\varphi$ spanning the manifold $M$ by leveraging the constraints inherent in rotative manipulators' \CS{}s. Adhering to the definition of the kinematic chain, the forward kinematic process follows a sequence of paired homogeneous transformations, where each rotation matrix depends solely on the last rotation matrix. This suggests that it is possible to plot \CS{} interactions of each dimension using only the last dependent coordinate while still representing the complete \CS{} and its dependencies. These dependencies are graphically depicted using colors and arc widths, combining elements of scatter plots, radial coordinates, hue, and saturation, as illustrated in \figref{fig:comparisonGS}.
\begin{figure}[tb]
\begin{center}
\includegraphics[width=0.98\linewidth]{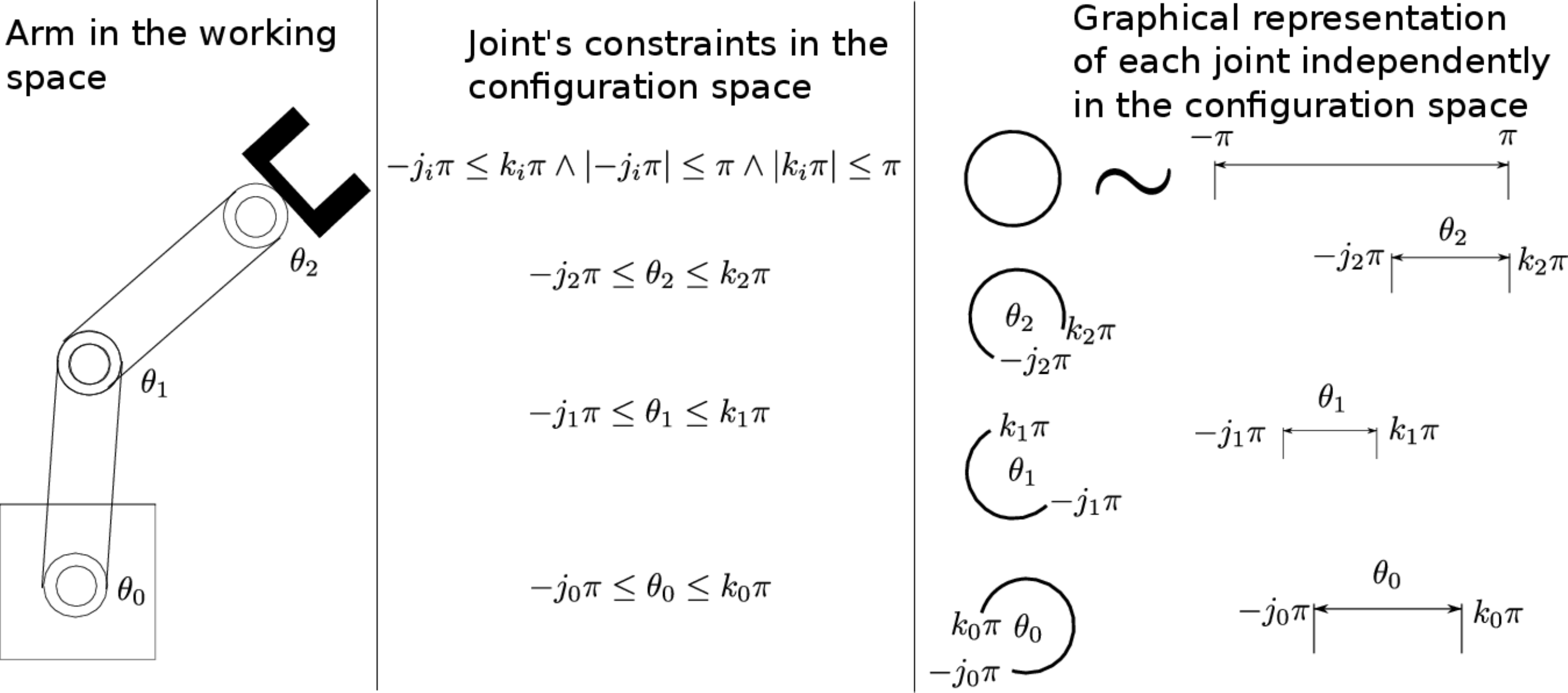}
\caption{A 3 DOF robot. We can easily observe the constraints of each joint of the robot by looking at the coordinate $\theta_i$ from each \CF{}-state $\bm{q}$, with $j_i,k_i \in \Real$ and  $i \in \NaturalNumbers$. Each unconstrained joint can be seen as a circle that is isomorphic to a line segment between $-\pi$ and $\pi$. It is not possible to observe the dependencies between the joints.}
\label{fig:armWithAngles}
\end{center}
\end{figure}

Our approach capitalizes on the isomorphism between a circle and a line segment to represent the angles within \CS{} as arcs or line segments. This strategy enables us to construct a visualization that accurately mirrors the rotations of each joint and the constraints governing them. Significantly, this visualization method possesses the distinct capability to highlight constraints as identifiable "holes" when no color points are plotted; within specific color dependencies. By utilizing this isomorphic technique, we establish an effective means of conveying the complex interplay between the robot's joints and their associated limitations.

\subsection{Kinematic Chain and Joint Interaction}
In the context of a rotative manipulator robot's \CS{}, characterized by the presence of $n$ joints, each state $\bm{q} \in$ \CS{} is represented as a tuple of $n$ scalars with $\bm{q}=[\theta_0,\theta_1,...,\theta_n]$, with $-\pi\leq\theta_i\leq\pi$ and $\Real^n,n \in \NaturalNumbers$. In the case of a specific sampled state \CS{} from a sampled dataset of size $m$; with $m\in \NaturalNumbers$, we would refer to the $j$th state as $\bm{q}_j$ with  $j \in \mathbb{N}, j \leq m$ and to the coordinates of such states as $\bm{q}_j=[\theta_0^j,\theta_1^j,...,\theta_n^j]$. Each coordinate $\theta_i^j$, can be independently visualized in a 2D projection. This visualization employs either polar coordinates in the form of $[\cos(\theta_i^j), \sin(\theta_i^j)]$ or a 1D bounded line segment within the subset of $[-\pi, \pi]$.

While visualizing $\bm{q}$ provides insight into the rotational boundaries of individual robot joints, it falls short in capturing the expressiveness of constraints imposed by the kinematic chain. For example, if $\theta_i$ is restricted to the interval $[-\pi, \pi]$ due to self-collisions or obstacles within the robot's WS, it has a cascading impact on joint $\theta_{i+1}$. However, when we independently plot all potential values of coordinate $\theta_{i+1}$ from a uniformly sampled dataset, it results in an almost continuous visualization of the joint. This continuous visualization obscures our ability to identify \CF{} regions shaped by constraints originating from $\theta_i$ and affecting $\theta_{i+1}$, as depicted in \figref{fig:armWithAngles}.

	To project multidimensional data into lower dimensions without losing the connection between non-consecutive entries of each data point, it is essential to maintain the relationship between the values within the $\bm{q}$ vector.

 An advantage of working with rotative manipulator robots is that each link connected to joint $\theta_{i+1}$ can be considered dependent only on the relative angle with respect to $\theta_i$, this can be seen as a series of 2D projections. The set of achievable states for the robot is defined by the values of each $\theta_i$, constrained by homogeneous transformations ${}^i_{i+1}\bm{T}(\theta_{i})$.
	\subsection{Combining  2D Projections and Parallel Coordinates}

We propose a solution to address the visualization of high-dimensional-data by 2D projections by combining labeling from parallel coordinates and the 2D projection of coordinate joints $\theta_i \times \theta_{i+1}$ by following the order of the homogeneous transformations of a manipulator robot. In this approach, each point in the 2D projection is labeled according to the order of the homogeneous transformations; $\theta_0^j, \theta_1^j, \ldots, \theta_n^j$.

	The integration of labeling enables us to preserve the relationship intra-joint of the original data from the visualization and simultaneously to identify gaps within each projection $\theta_i \times \theta_{i+1}$ as the regions where clusters of points are not uniformly distributed or empty. %

Since we have a discrete number of samples, interpreting the impact of each individual point on the local regions becomes challenging; depending on the spread of the samples, the data points may not effectively convey the topology of \CS{}.

Now that we know that it is feasible to keep most of the n-dimensional \CS{} when building the 2D data structures, the act of visualizing continuous data to a discrete format for the practitioner would result in the loss of information; this happens even when the visualization has the same dimension as the data being plotted. The discretization also creates branching of the values of the original data. This loss of information can be used in our advantage to plot discretization clusters that approximate of the original data by its Euclidean distance with a sequence of 2D projections.

The visualization of $\theta_i$ given $\theta_{0},...,\theta_{n}$ will be subject to discretization when viewed on an output device; which impose limitations on the number of pixels or file size represented as discrete grid. Consequently, the potential for displaying various values for each pair of $\theta_i$ and $\theta_{i+1}$ will be constrained by the discretized state samples. This is achieved through $\hat{\bm{q}}_j = \bm{q}_j + \bm{\sigma}$, where $\bm{\sigma} = [\sigma_0, ..., \sigma_n]$ and $\sigma_i \in \Real$.

\begin{figure}
	\begin{subfigure}{\linewidth}
		\centering
	    \includegraphics[width=0.7\columnwidth]{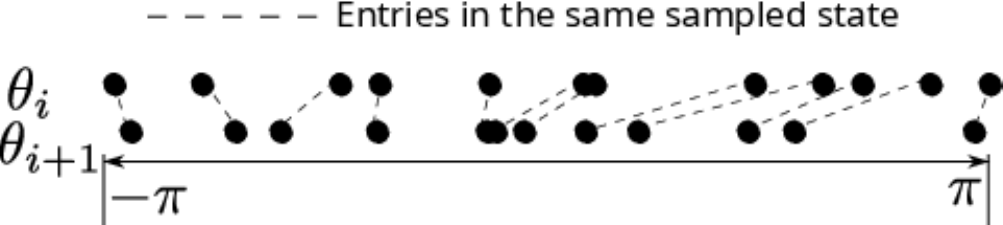}
	    \caption{Each time that we sample a \CF{}-state $\bm{q}_j$, we consider that each coordinate $\theta_i^j$ has a relationship with the rest of the entries $\theta_k^j$ of the $j$th sample. This relationship can be ordered following the kinematic chain from $\theta_0$ to $\theta_n$.}
	     \label{fig:discretization1}
	\end{subfigure}
		\begin{subfigure}{\linewidth}
			\centering
	    \includegraphics[width=0.7\columnwidth]{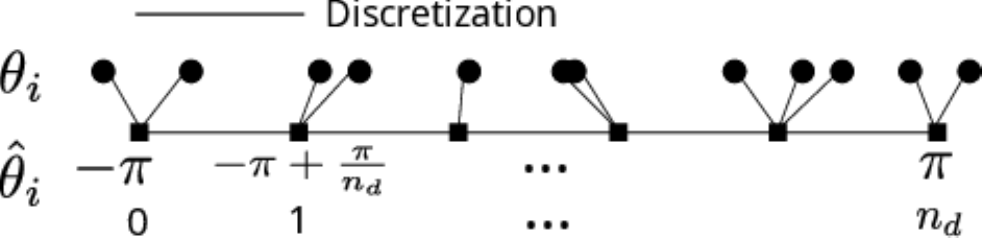}
	    \caption{The visualization from \CS{} will be represented in a discrete output, each $\theta_i^j$ will be approximated given the resolution of the output device. We can control the discretization before being sent to the output for visualization by distributing uniformly the sampled data given a uniform interval of size $\frac{1}{n_d}$ where $n_d \in \NaturalNumbers$.}
	     \label{fig:discretization2}
	\end{subfigure}
			\begin{subfigure}{\linewidth}
				\centering
	    \includegraphics[width=0.7\columnwidth]{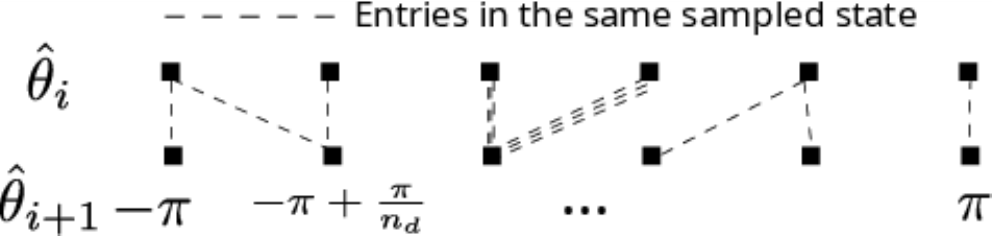}
	    \caption{The new vector states $\hat{\bm{q}}_j=\bm{q}_j+\bm{\sigma}$ will be created by keeping the relationship of the previous $\theta_i^j,\theta_{i+1}^j$ on the perturbed by discretization values $\hat{\theta}_i^j,\hat{\theta}_{i+1}^j$.}
	     \label{fig:discretization3}
	\end{subfigure}
	\caption{Discretization effect when the continuous values from \CS{} are displayed on a discrete output device.}
	\label{fig:discretization}
\end{figure}

The visualized $\hat{\bm{q}}_j$, will increase the probability of observing $\hat{\bm{q}}_j=\hat{\bm{q}}_k$, where $k \leq m$ and $k \neq j$ as the sample size grows, provided the discrete resolution remains constant. This is visually depicted in \figref{fig:discretization}. The discretization naturally leads to a tree-like structure, reflecting the interdependence of individual joints. The children of the roots $\theta_i$ are expressed as follows:
{
\begin{multline}
    \theta_{i+1}|\theta_i^j=\{\forall\theta_{i+1}^k: \theta_{i+1}^k\in \bm{q}_k \land  \theta_i^k=\theta_i^j, \\
    \forall k \in \NaturalNumbers,0\leq k\leq m \}
    \label{eq:tree}
\end{multline}}

For the sake of generality, we will treat $\bm{q}$ as equal to $\hat{\bm{q}}$, considering that $\bm{q}$ is sampled from a uniform distribution. As the number of discretization partitions in the uniform interval $n_d \in \NaturalNumbers$ approaches infinity, we have $\bm{q} \approx \hat{\bm{q}}$.

This tree-like structure approach enables the detection of patterns within the sampled points cloud of states by considering the interdependencies between the connected joints and allow us to detect collision even in high-dimensional \CS{}s by following the kinematic chain of the robot in 2D projections.

\section{Visualization}\label{sec:visualization}
Now that the data is organized into 2D sequenced projections, our focus shifts to effectively conveying this information to the user. This section aims to provide a clear visualization that enables users to discern \CF{} regions and identify the constraints within \CS{}.

\subsection{Keeping the Index of the Sampled States in the Visualization}

A key challenge is to uniquely and graphically identify the root values expressed by the trees $\theta_i|\theta_{i+1}$ without inflating the dimensions of the visualization and also preserving the information of each independent sampled state. One elegant solution to address this challenge is to employ color mapping.

Color mapping is a powerful technique that can be likened to a curve within the 3D color space \cite{Kovesi2015GoodCM}. In our approach, we employ color to represent the Euclidean distance within \CS{}. Specifically, states that are close in terms of their parent values and current configurations will exhibit similar colors in the visualization. This color mapping not only aids in visually capturing \CS{} structure but also helps practitioners in clustering states within \CF{} local regions.

Importantly, we maintain the visualization in 2D, as research has shown that 2D visualizations can outperform their 3D counterparts, particularly in tasks related to obtaining an overview of the data \cite{10.1145/3313831.3376675}.

The color mapping strategy involves associating a distinct color with each root tree $\theta_{i+1}|\theta_i^j$. The creation of these trees is necessary because of the limited color depth available in the output device, which constrains the number of colors that can be effectively represented. Although it is possible to assign a unique real number to each state in the color map, the mapping from the color map to RGB values would result in very close values in the color map domain being assigned to the same RGB color. From the user's perspective, two states with nearly identical RGB colors would appear the same. This process can be likened to discretizing the original data, leading to the formation of a tree-like structure as described in Eq. \eqref{eq:tree}. Given that each tree can be viewed as a 2D division of the $n \times m$ \CS{} grid, it becomes feasible to implement a unique color map for each joint $\theta_i$.

The process begins by dividing the color mapping across the uniform interval of $[-\pi, \pi]$ and the discretization size $n_d$ for each coordinate $i$. In practical terms, when we have a parent value $\theta_i^j$, we select a specific color from the color map. This chosen color is then assigned to all the children $\theta_{i+1}|\theta_{i}^j$ of the parent value, creating a clear visualization of the kinematic constraints within \CS{}. The color assignment process is shown in  \figref{fig:circleGeneration}.

In order to preserve as much of the information between non-consecutive joints of the original dataset encoded by the relationship between the 2D trees, $\theta_{i+1}|\theta_i$ and the rest of the joints in the sampled state, each time we assign colors to trees of the form $\theta_{i+1}|\theta_i^j$; where $\theta_{i+1}^j$ can be also a child of different parent $\theta_i^k$, we add a slightly different color to the states $\bm{q}_k$ by a perturbation $\bm{\epsilon}_k=\epsilon_k[1,...,1]^T \in \mathbb{R}^n,$ $\epsilon_k  \in \Real^+$. While introducing noise to the original data may introduce imprecision to the original model, it can help prevent the unbounded error associated with the possible representation of additional states resulting from the branching of the data. It is worth noting that the error of the perturbed state is bounded by $\|\bm{\epsilon}_k\|$ given that the perturbation follows $\|\bm{q}_k+\bm{\epsilon}_k\|\leq \| \bm{q}_k\|+\|\bm{\epsilon}_k\|$.
\subsection{Visualizing the States in $S^1$ }
After assigning the colors to the states, we proceed to create a visualization within the context of a rotative manipulator robot, we have adopted a method that projects each tree onto a segment of the circumference of $S^1$. This approach allows us to effectively identify the angles $\theta_{i+1}$ at which the joint is in a \CF{} state. Much like the tree-based visualization, the color assigned to each arc corresponds to the angle of the parent joint, which is denoted as $\theta_i$. Essentially, each parent $\theta_i^j$ is visually represented by a dedicated arc on the circumference of $S^1$.
\begin{figure}
\begin{center}
\includegraphics[scale=0.17]{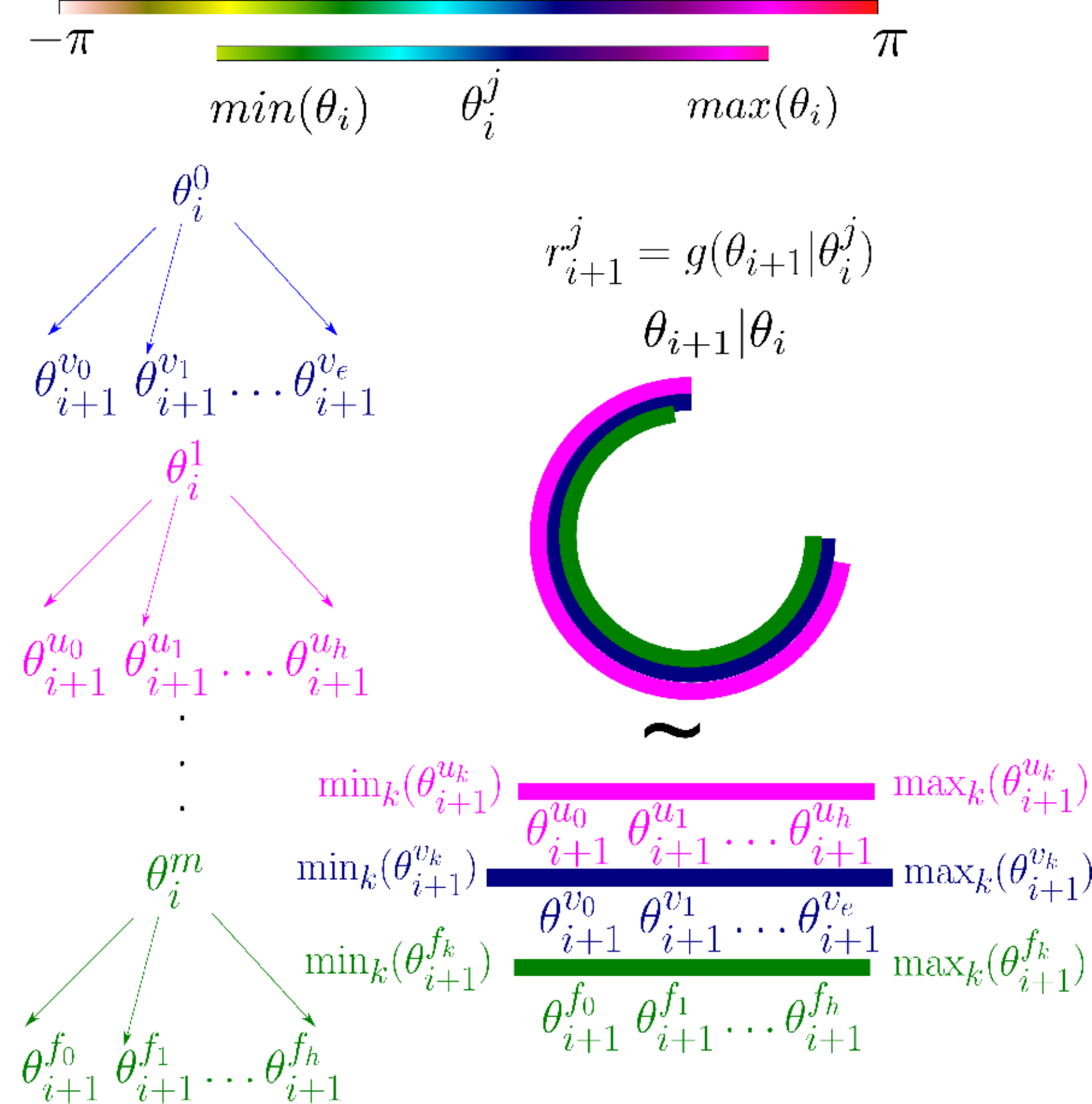}
\caption{Each parent joint $\theta_{i}$ can be ordered given the interval $[-\pi,\pi]$. Then we assign a color to each tree $\theta_{i+1}|\theta_{i}^j$ by distributing uniformly the value of the joint $\theta_i^j$ given the number of samples in the color space. All the colored children from $\theta_{i}|\theta_{i+1}^j$ are assigned a radius $r_{i+1}^j$ by an increasing function $g(\theta_{i+1}|\theta_i^j)$ and are plotted in their respective 2D positions $g(\theta_{i+1}|\theta_i^j)cos(\theta_{i+1}^j),g(\theta_{i+1}|\theta_i^j)sin(\theta_{i+1}^j)$. The length of each circumference is given by the maximum and minimum $\theta_i$ from each $\theta_{i+1}$}
\label{fig:circleGeneration}
\end{center}
\end{figure}
\begin{figure}[ht]
\begin{center}
\includegraphics[scale=0.10]{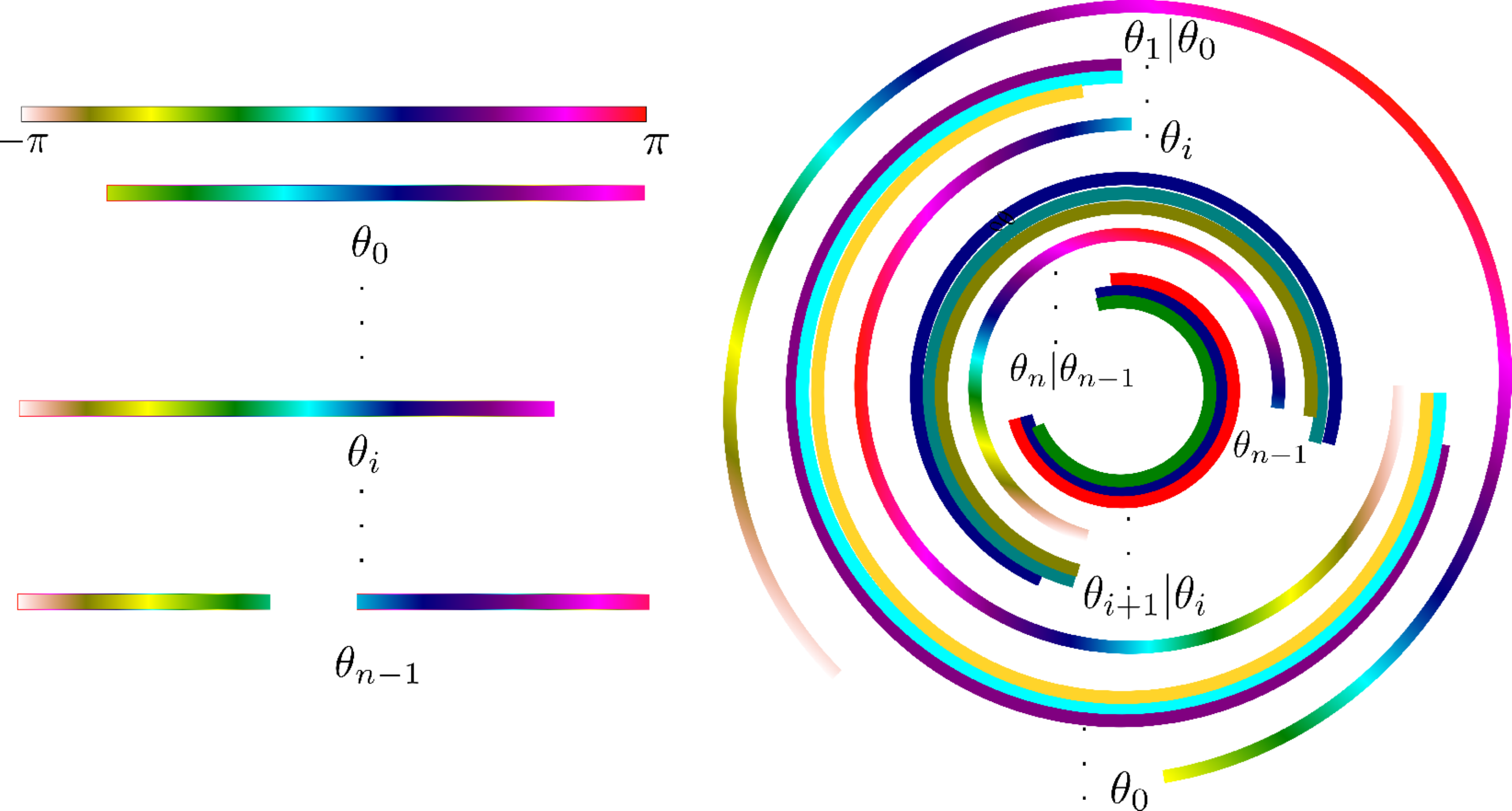}
\caption{The color map arc $\theta_i$ represents all the possible entries of the joint given the dataset. This helps us to localize the original values of the sampled states in radians following the kinematic chain of the robot. 	With this color assignment, we are able to provide a guideline to create a visualization of the distribution of \CF{} states in high-dimensional \CS{}s without losing the initial indexing of the data samples.}
\label{fig:unionOfGraphsPlusGuides}
\end{center}
\end{figure}
In order to visualize all the colored arcs $\theta_{i+1}|\theta_{i}^j$ distinctly, we employ an increasing function denoted as $g(\theta_{i+1}|\theta_{i}^j)$ to assign a radius $r_{i+1}^j$ to each joint $i+1$ and sample $j$. This technique ensures that the arcs do not overlap with each other. The length of each arc is determined by the specific kinematic constraints of the joint. The reader can refer to \figref{fig:circleGeneration} for a visualization of this process.

By appropriately scaling each set of arcs $\theta_{i+1}|\theta_{i}^j$, it becomes possible to plot the values of all the joints simultaneously. %

To easily indicate the relationship between the colors and the corresponding angle of the parent joint $\theta_i$, a color legend is employed. This legend visually associates colors with their specific angle in the parent joint $\theta_i$.

To further improve clarity and readability, we add a chart $\varphi$ that connects the coordinate $\theta_i$ and $\theta_{i+1}$ in the form of a line segment spanning the interval $[-\pi,\pi]$. This line segment defines the color map for the trees/arcs of the children $\theta_{i+1}|\theta_i$ and represents the boundaries of the coordinate $\theta_{i}$, independent of any other coordinate $\theta_{i+k}, k \in \mathbb{Z},  k \neq 0$. The visualization of the process is illustrated in \figref{fig:unionOfGraphsPlusGuides}.
\section{Visualization Results}\label{sec:representation}

In this section, we present the key findings derived from our analysis both qualitatively and quantitatively. We illustrate the improvement of the training of a neural network model over several epochs, demonstrating how this visualization gradually becomes similar to the original \CS{} data. Additionally, we provide a quantitative analysis to support the claim that the information conveyed by the 2D visualization accurately reflects the improvements in accuracy measurements, even for \CS{}s with more than three dimensions. This analysis is particularly noteworthy in cases where only a subset of \CS{} is being represented. The source code is available on \footnote{https://bitbucket.org/joro3001/multidimensionalplots/}.

	\subsection{Qualitative Evaluation}\label{sec:qualitative}
\begin{figure}[ht]
\centering
\begin{subfigure}[t]{.12\textwidth}
	    \includegraphics[scale=0.14]{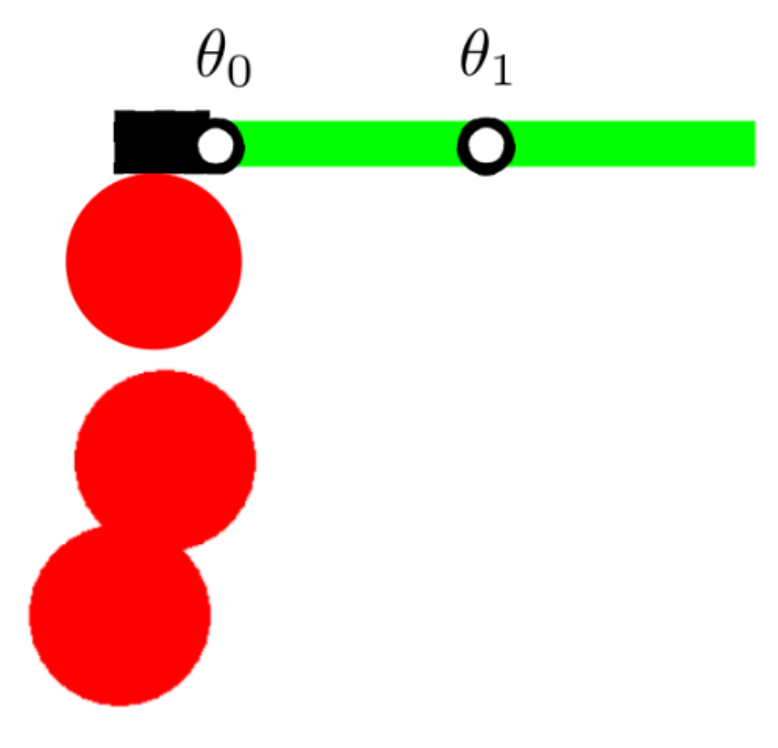}
	    \caption{2D robot. The links of the robot are the green rectangles. The red circles are obstacles in the WS.}
	     \label{fig:ws2dwithJoint}
	\end{subfigure}\hspace{0.01\textwidth}
		\begin{subfigure}[t]{.15\textwidth}
	    \includegraphics[scale=0.11]{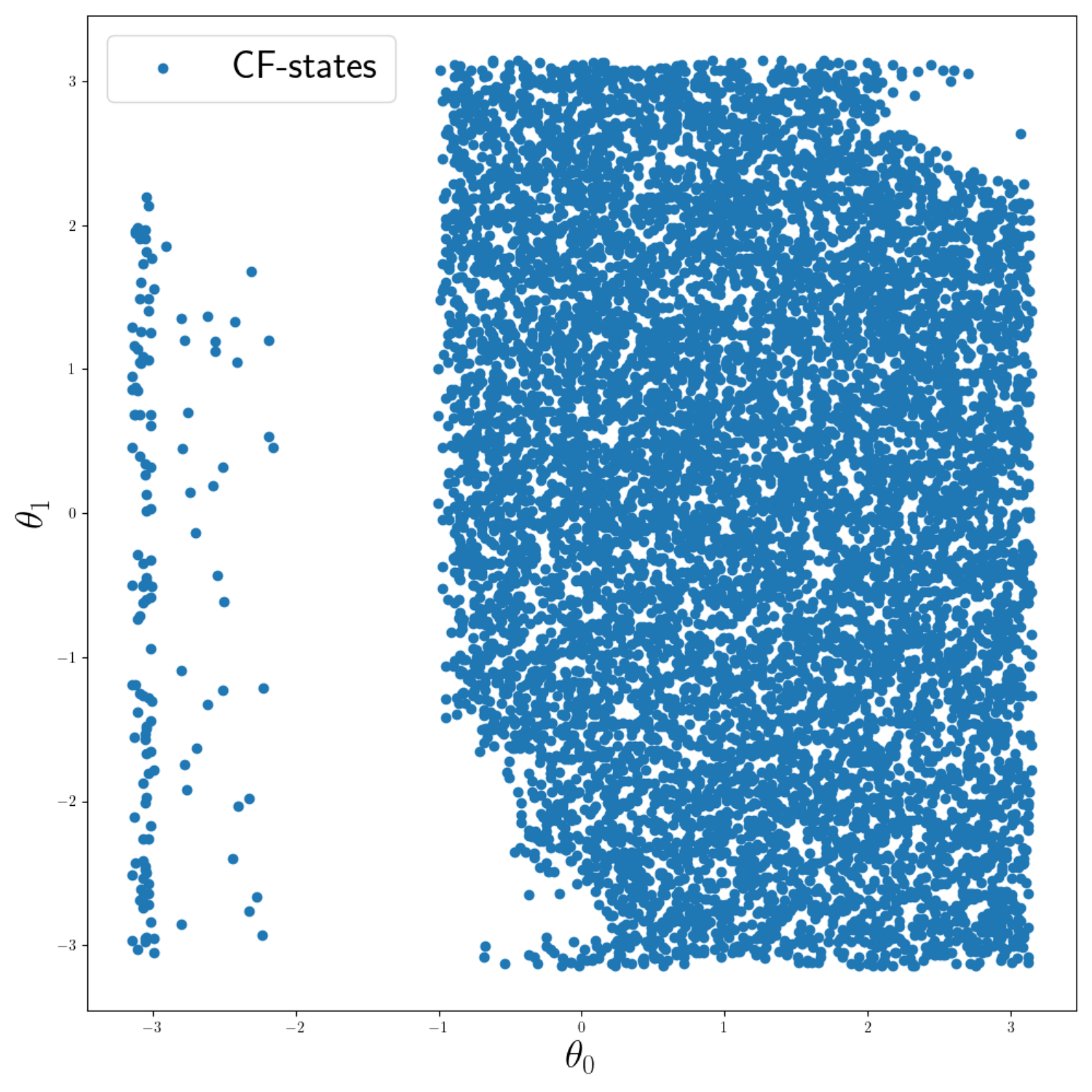}
	    \caption{$\theta_0\times \theta_1$ projection of the 2D \CS{} of the robot in radians.}
	     \label{fig:cs2dProjection}
	\end{subfigure}\hspace{0.01\textwidth}
			\begin{subfigure}[t]{.16\textwidth}
	    \includegraphics[scale=0.12]{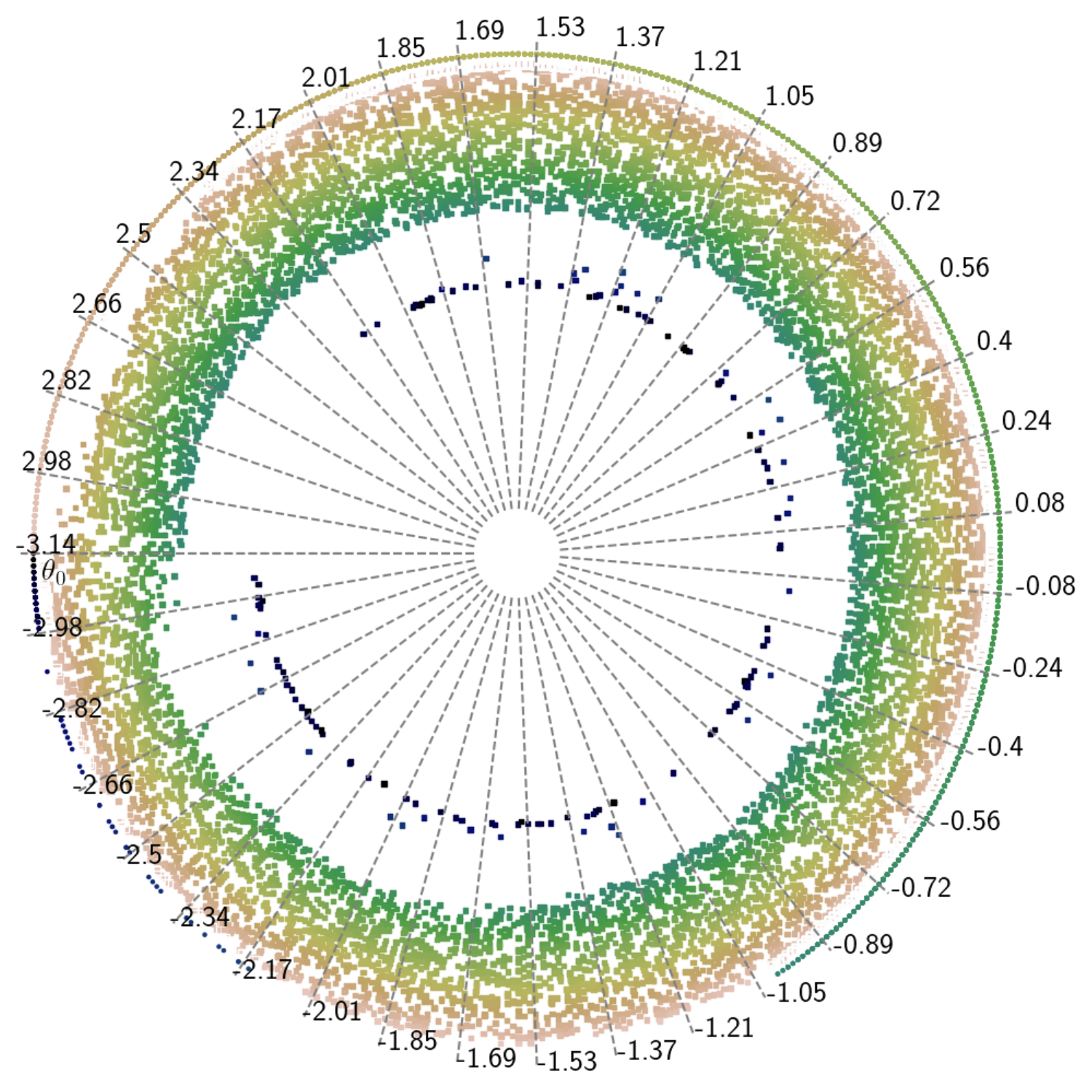}
	    \caption{Proposed visualization. The gap of the $\theta_0\times \theta_1$ projection is shown as the lack of points with colors in $\theta_1|\theta_0$ following the gradient of the map $\theta_0$}
	     \label{fig:wholeGuides2d}
	\end{subfigure}
	\caption{WS and \CF{} simulation of 2 DOF. Our visualization highlights the changes that come from the dependency of the two joints: the lack of colors in the 2D arc $\theta_1|\theta_0$ means that the collision states were generated by constraints in $\theta_0$, the lack of continuity of points of the same colors indicates constraints coming from the joint $\theta_1$.}
	\label{fig:2dExample}
\end{figure}

In the context of a simple 2D manipulator robot, our proposed visualization naturally captures the essence of the 2D projection of $\theta_0\times \theta_{1}$ as depicted in the trivial example in \figref{fig:2dExample}. The challenge is to demonstrate how effectively this concept translates to high-dimensional \CS{}s.

To evaluate the effectiveness of our visualization in high-dimensional \CS{}s, we gathered a dataset of \CF{} states from \CS{} of a 7-DOF right arm of a Baxter industrial robot. This evaluation included scenarios where a simulated human presence within the robot's WS introduced geometric constraints, which significantly impacted the visualization of \CF{}, as illustrated in \figref{fig:comparisonGS}.

To estimate collisions between the robot and the simulated human and construct a training dataset, we utilized MoveIt \cite{barrier2014} for simulating and collision checking.

We changed the position and orientation of the human in two different random configurations within the WS of the robot. We sampled 10,000 points for each configuration to construct \CS{}.

We proceeded to train a small example to assess the capability of representing high-dimensional-data, with the objective of comparing how well a trained model can approximate the distribution of the original \CS{}.

We trained a Generative Adversarial Network (GAN) \cite{9385935} to develop a model that can generate samples from one of the two configurations. We employed the architecture proposed in \cite{jimenez2023improving}, with the modification that the WS condition is now incorporated by simply concatenating 0 or 1 to the input of the latent vector.

We selected the ``GistEarth" color map for mapping $\theta_i$ and $\theta_{i+1}|\theta_i$. This choice was influenced by the resemblance of this color map to luminance-controlled maps, such as ``Rainforest" \cite{Velden_2019}. Luminance is a critical factor as it conveys information about the structural aspects of surfaces in three-dimensional space \cite{Marr:1982:VCI:1095712}. Moreover, this color map remains interpretable in grayscale for individuals with color vision deficiencies \cite{Velden_2019}.

\begin{figure}[ht]
\centering
\begin{subfigure}[t]{.15\textwidth}
	\includegraphics[width=1\linewidth]{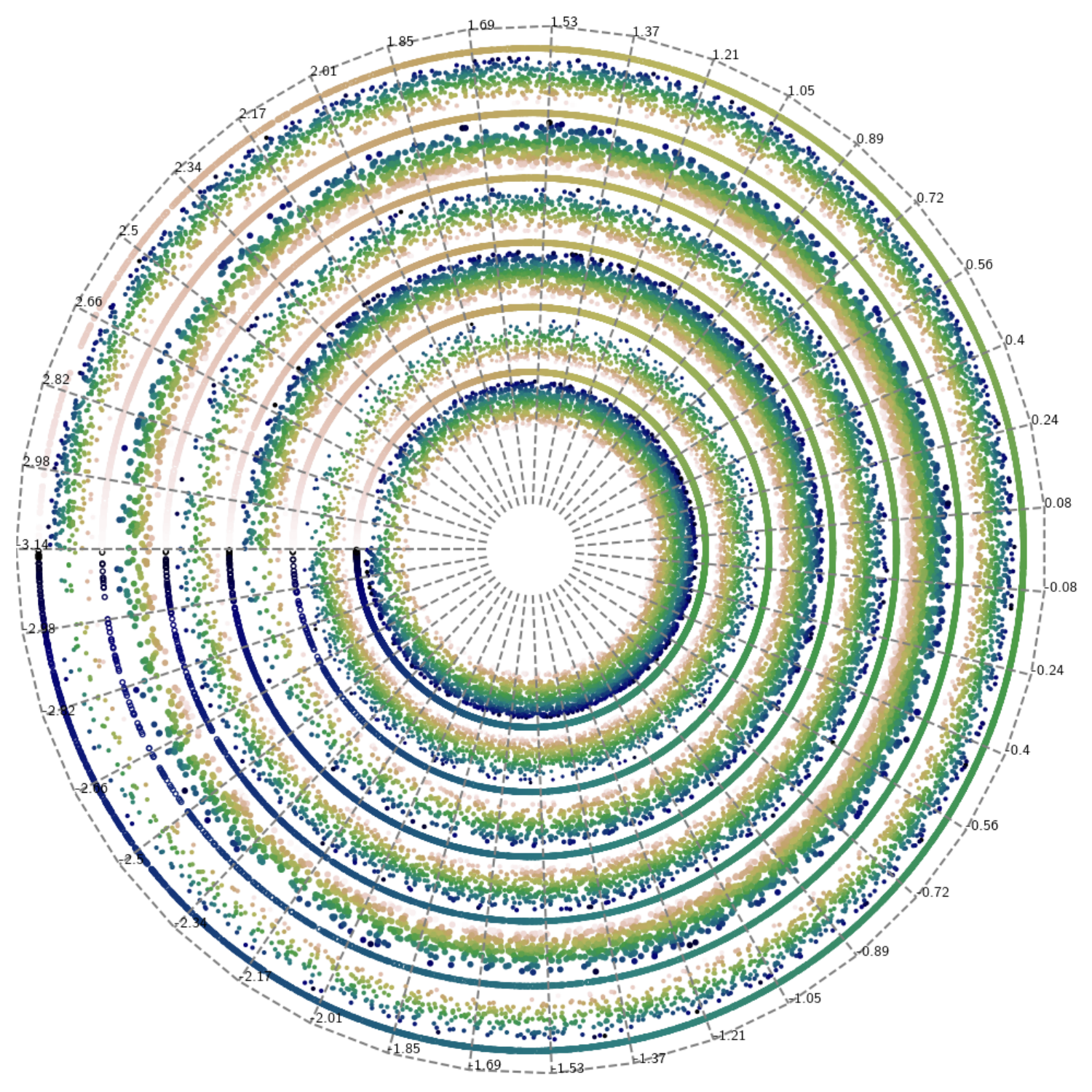}
	      \caption{0 epochs}
	      \label{fig:0trainedModel}
\end{subfigure}
\begin{subfigure}[t]{.15\textwidth}
	 \includegraphics[width=1\linewidth]{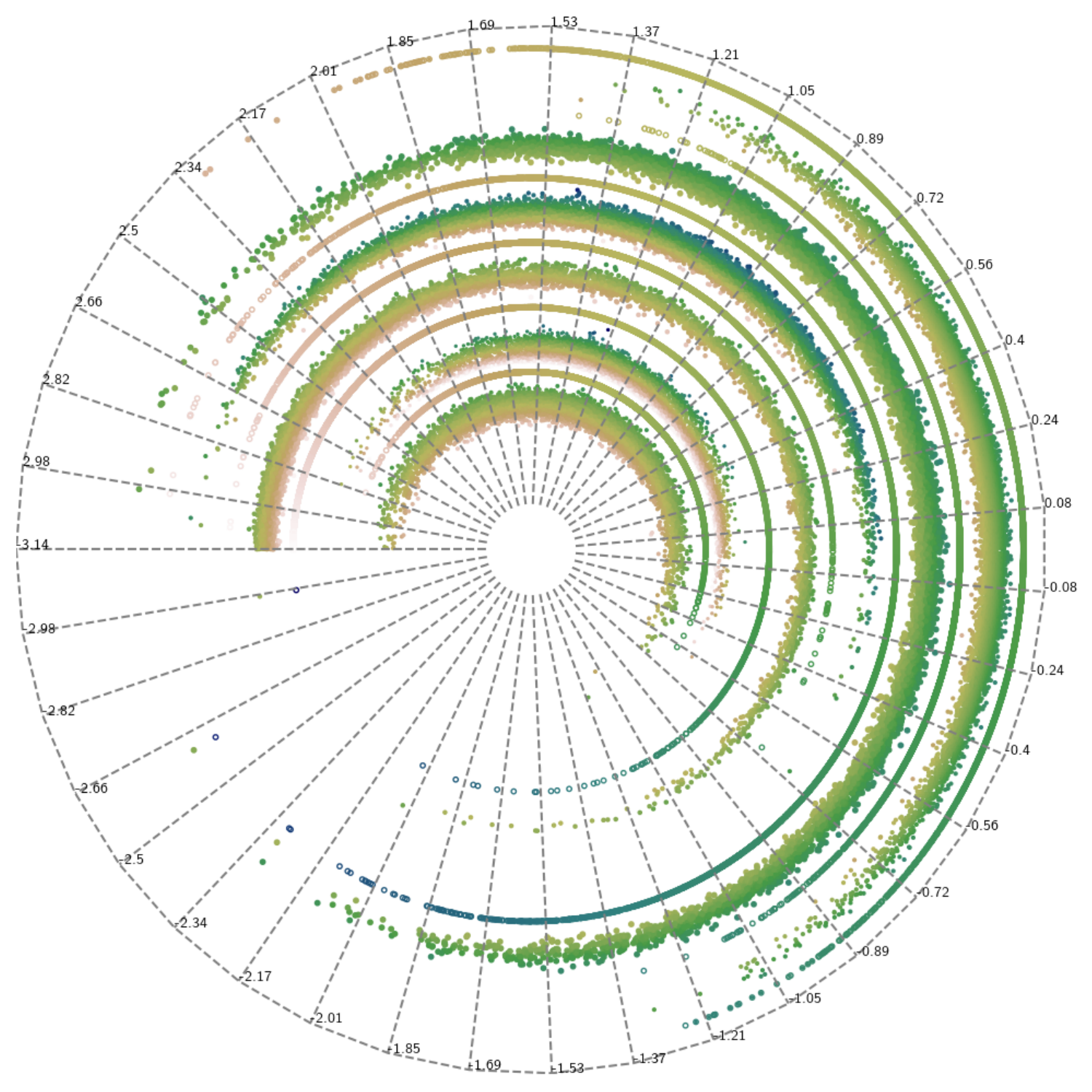}
	      \caption{10 epochs}
	      \label{fig:10trainedModel}
\end{subfigure}
\begin{subfigure}[t]{.15\textwidth}
	 \includegraphics[width=1\linewidth]{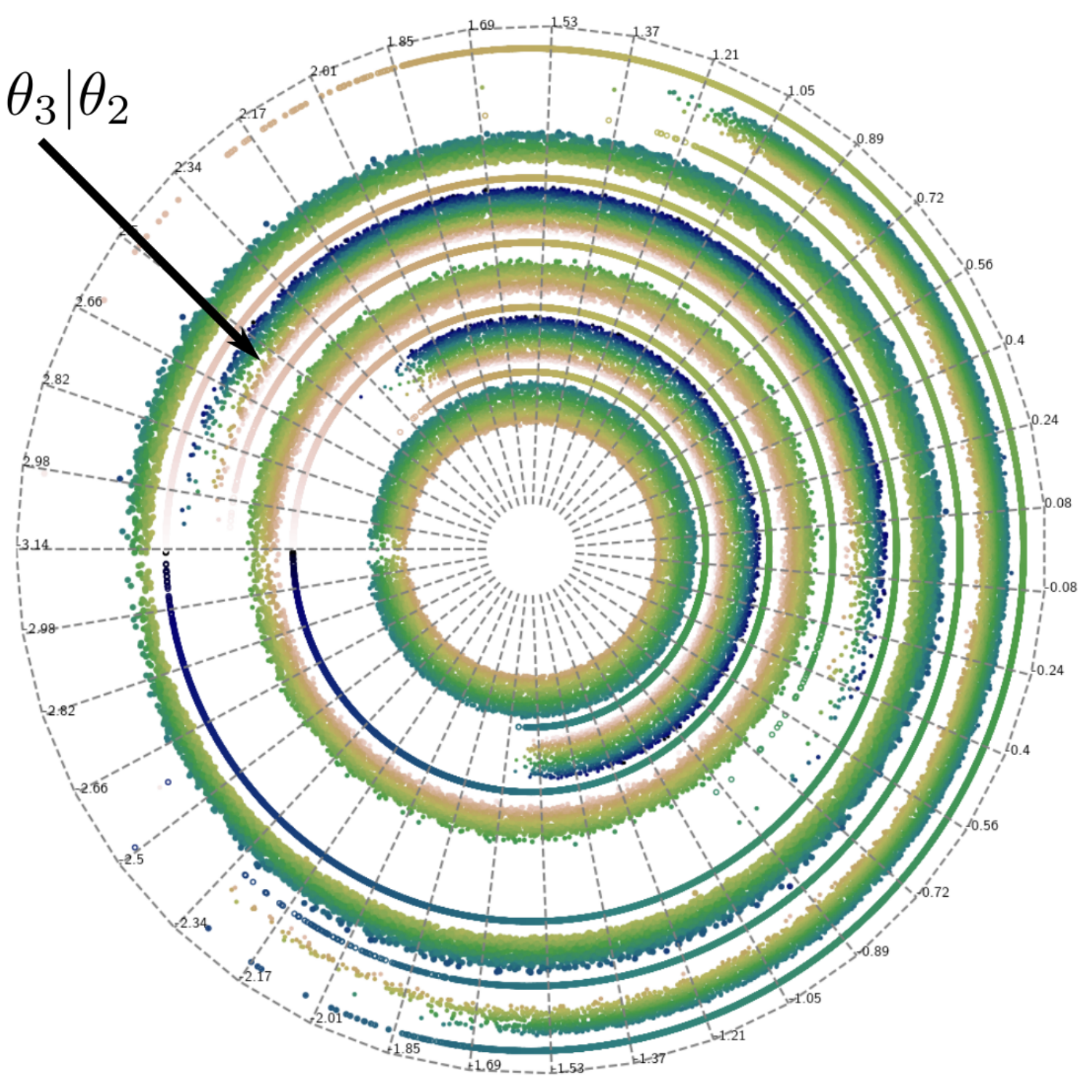}
	      \caption{100 epochs}
	      \label{fig:trainedModel-1_0.png}
\end{subfigure}
\caption{Representation of the original \CS{} of 7 DOF from \figref{fig:comparisonGS} by training a GAN. We can see that given the increasing of epochs the model starts to reproduce more closely the training data.}
\label{fig:evolutionTraining}
\end{figure}
\begin{figure}[!ht]
\centering
\setlength{\fboxsep}{0pt}
	\setlength{\fboxrule}{0pt}%
\fbox{
\begin{subfigure}[t]{.25\linewidth}
	\includegraphics[scale=0.1]{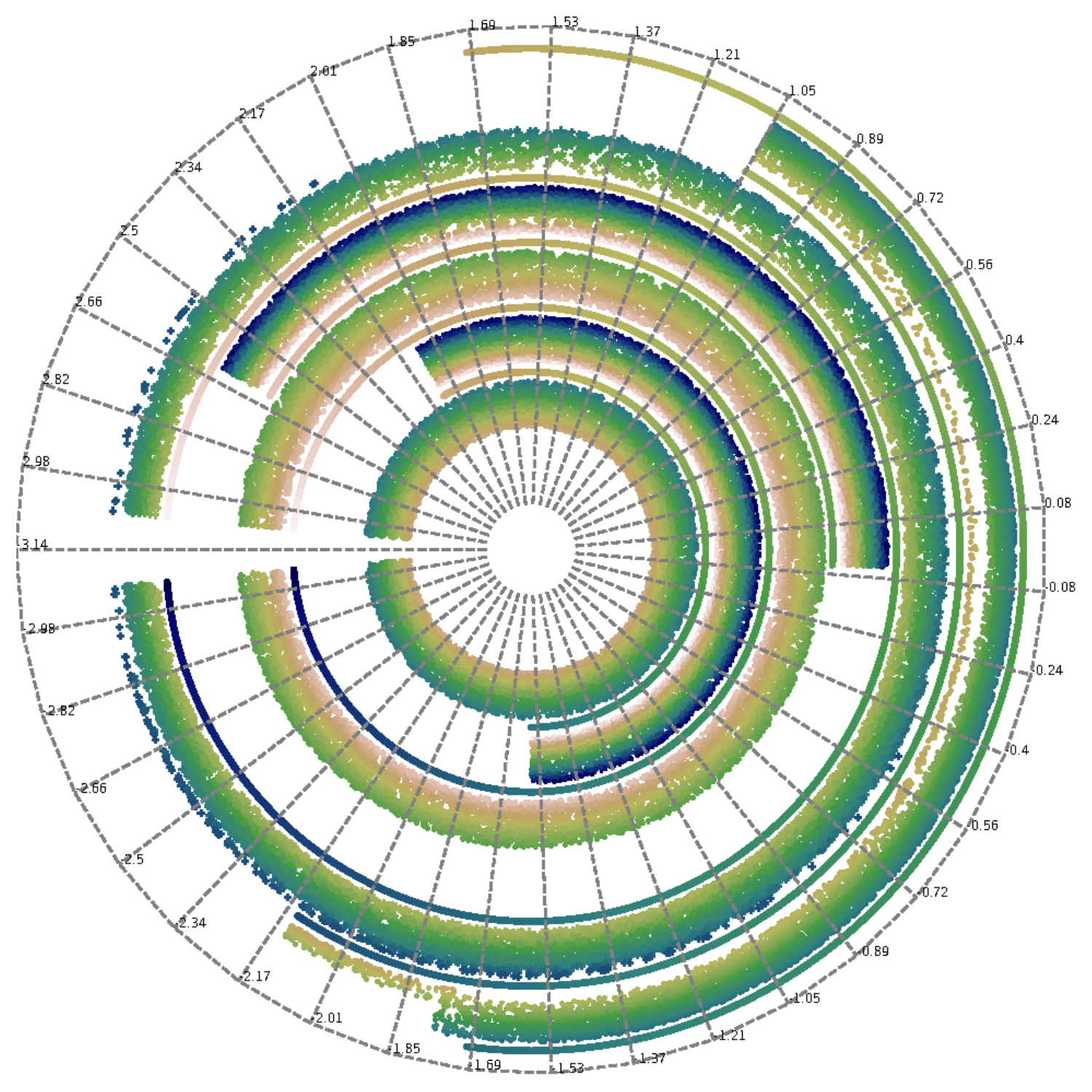}
	      \caption{CS1}
	      \label{fig:cs1}
\end{subfigure}
}
\fbox{
\begin{subfigure}[t]{.25\linewidth}
	 \includegraphics[scale=0.1]{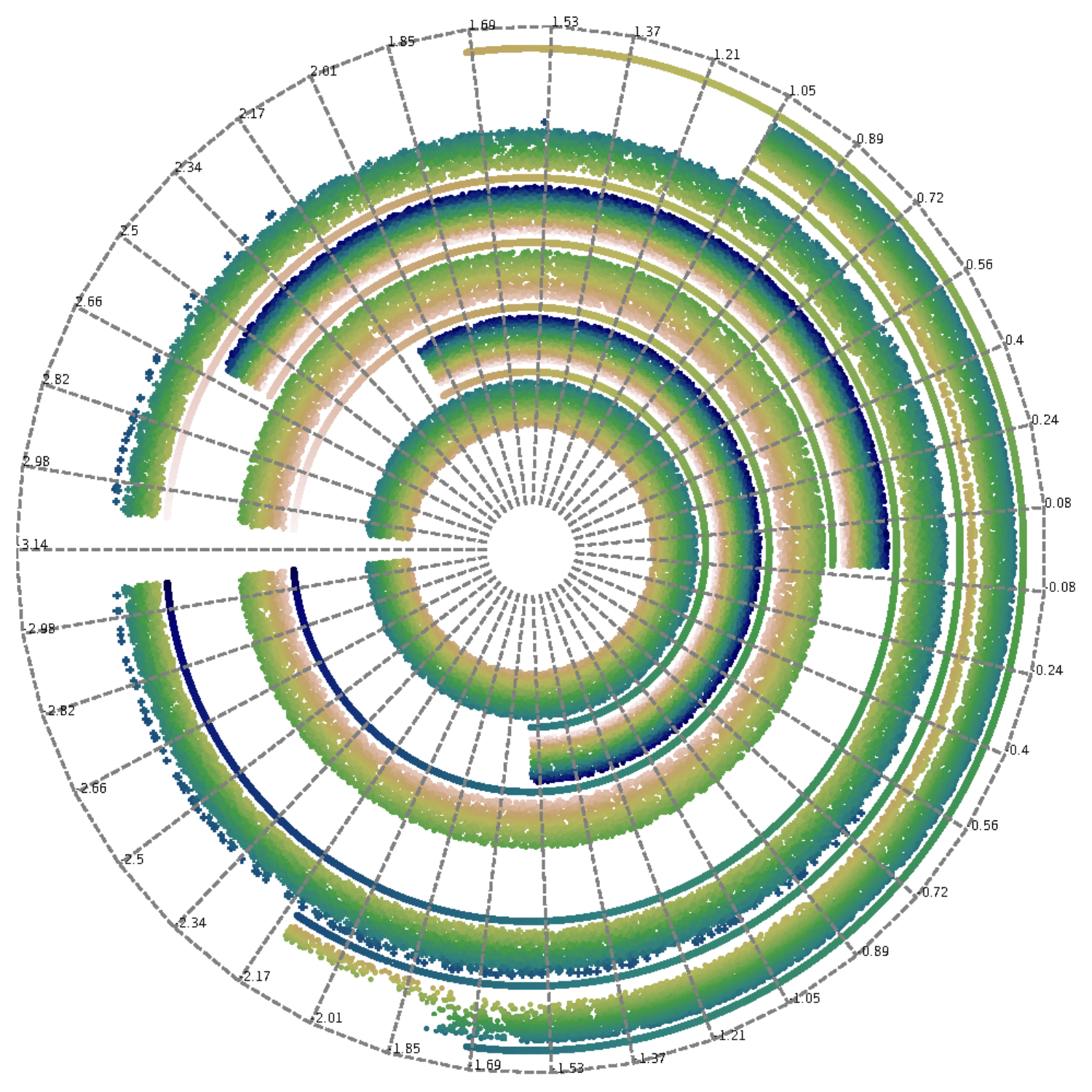}
	      \caption{CS2}
	      \label{fig:cs2}
\end{subfigure}
}
\fbox{
\begin{subfigure}[t]{.4\linewidth}
	 \includegraphics[scale=0.08]{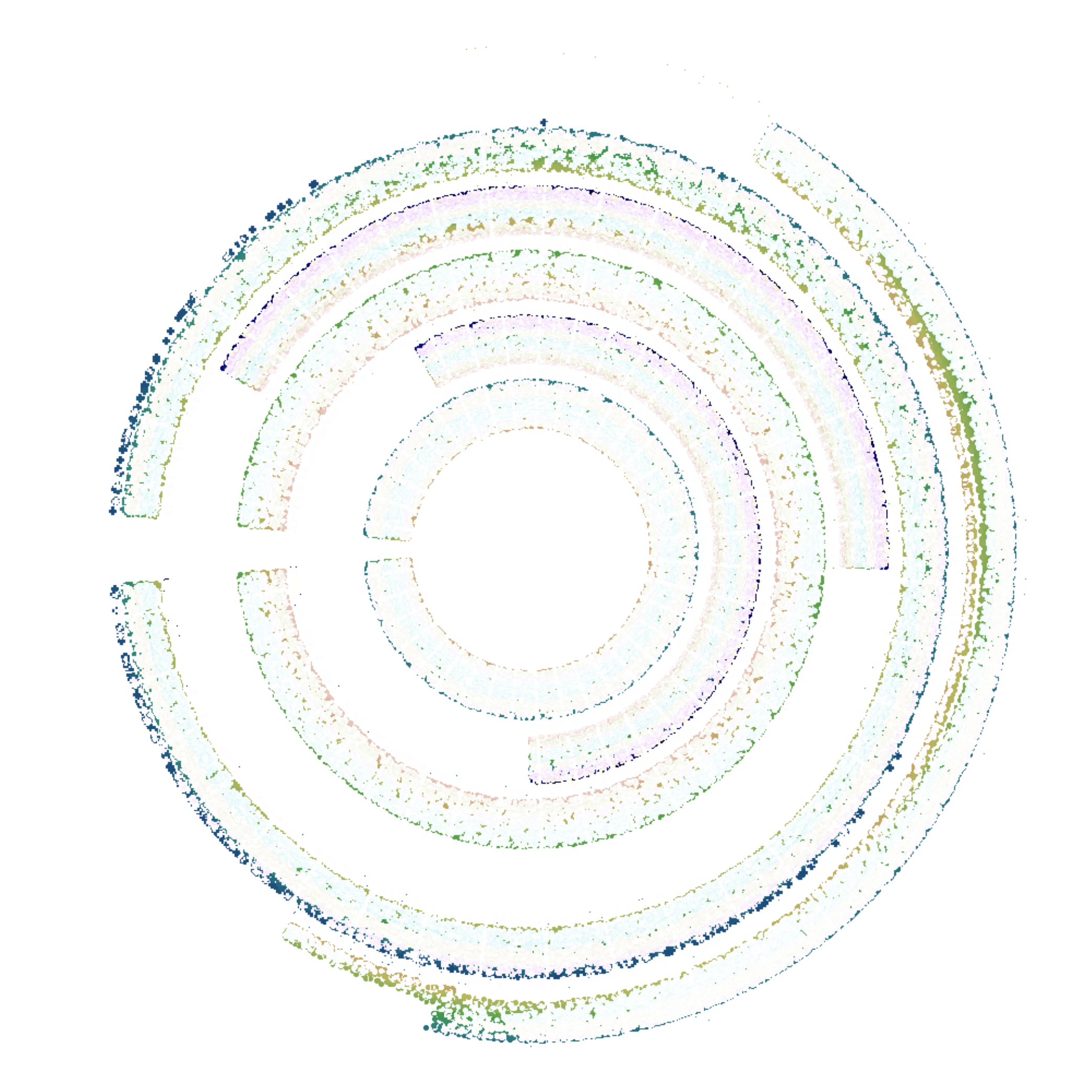}
	      \caption{Negative image of the subtraction of CS1-CS2}
	      \label{fig:substractionNegative}
\end{subfigure}}
\caption{\CS{}s that look similar when the WS changes. We can use the difference between \CS{}s visualizations to localize more easily the regions were \CS{}s diverge. This approach is not available with other high-dimensional visualization techniques given the occlusion of the sampling points or the lack of relationships between all the joints while using a lower-dimensional projection. }
\label{fig:comparisonCSs}
\end{figure}

As demonstrated in \figref{fig:evolutionTraining}, our visualization effectively illustrates the model's progress in learning the original \CS{} as training epochs increase. Notably, the model continues to refine its understanding of the distribution governed by its parent joints. For example, the arc $\theta_3|\theta_2$ illustrates this ongoing learning process, as it has yet to fully capture the interrelationships between the joints, resulting in the distribution of colors in the visualization. This is further evidenced by the multitude of small holes in the original data, depicted in \figref{fig:comparisonGS}. These gaps denote potential collision configurations that the model is still in the process of learning them.

Upon closer examination, we observed that when two \CS{}s exhibit variations in only limited regions, it can be challenging for users to pinpoint these specific alterations within \CS{}. To address this issue, we leverage our image visualization to identify such areas of interest. By employing image subtraction techniques, we can effectively extract regions where differences between two \CS{}s are localized. These areas of interest are indicated to the user, guiding them to investigate specific regions exhibiting dissimilarity, highlighted by the values of the pixels of darker colors. This approach is exemplified in \figref{fig:comparisonCSs}.

Employing this approach, our proposed visualization effectively identifies disparities between two similar yet distinct \CS{}s. By highlighting only the regions where the distributions diverge and by adjusting color intensity, we can gauge the proximity of the \CSA{} to the original \CS{} representation.
\subsection{Quantitative Evaluation}\label{sec:quantitative}
In the field of robotics SBMP, the evaluation of \CS{} representations typically occurs when a machine learning model is put into action in an actual motion planning task. The primary aim would be to minimize the need for extensive collision-checks, thereby accelerating the search for \CF{}-paths. The evaluation of these paths predominantly revolves around two key metrics: the number of \CF{}-paths discovered and the query time. However, this approach comes with a notable challenge.

\begin{figure}[h]
\begin{center}
\includegraphics[scale=0.08]{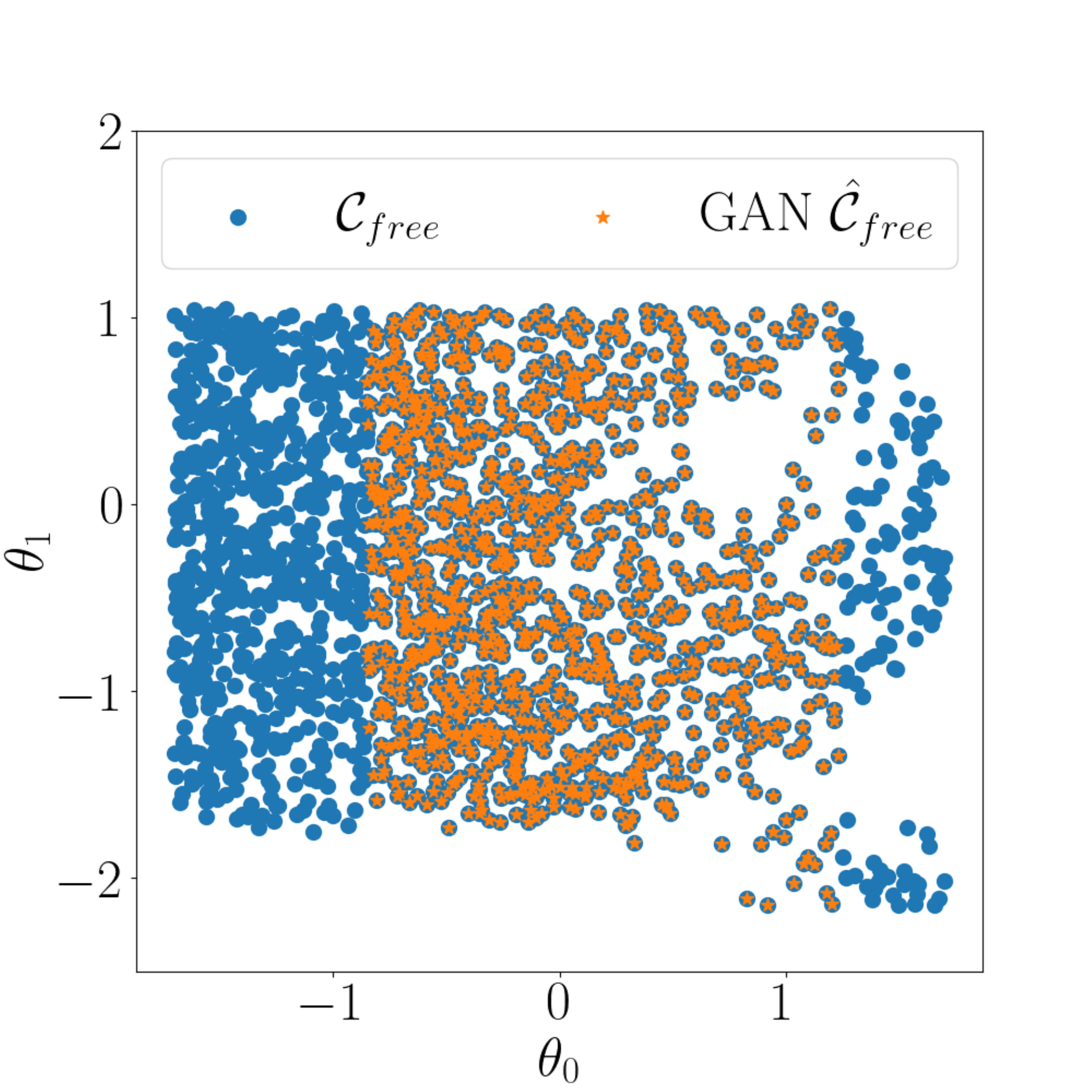}
\caption{Mode collapse in GAN:  The generator covers a small subset of the true \CF{}; the collision checker achieves high accuracy when assessing the generated samples.}

\label{fig:modeCollapse}
\end{center}
\end{figure}

For instance, when a Generative Adversarial Network (GAN) is employed to guide a sampler, the issue of mode collapse can arise. Mode collapse is a phenomenon where the generator predominantly produces a single sample or a very limited set of closely similar samples \cite{metz2017unrolled}, thus ignoring a significant portion of \CS{}. As a result, when the robot is deployed in real-world scenarios beyond the boundaries of the GAN's collapsed representation, it may miss crucial \CF{}-regions or paths. This situation is analogous to relying solely on accuracy and collision checking to measure a subset of \CF{}. An illustrative example of this phenomenon is depicted in \figref{fig:modeCollapse}.

Although our visualization offers valuable qualitative insights into high-dimensional \CS{}s, it also has the potential to assess cases where only subsets of \CS{} are generated. Furthermore, it can serve as an approximation tool, given that it displays 2D projections of data from each pair of joints, $\theta_i$ and $\theta_{i+1}$, along with a chart, $\varphi$.

First, we conducted a test of our visualization to determine if the visualization is able to capture information about accuracy from the collision checker. In this context, accuracy is defined as the ratio of \CF{}-samples generated by the trained sampler, validated by the collision checker, to the total number of samples queried from the trained model.

We used the 100 different WS/\CS{} pairs from the previous section. In these experiments, we introduced random collision states inside the boundaries of the joints and replacing some of \CF{}-states of the original \CS{}. The amount of added collision states went from 10\% to 100\% of the total size of samples of each \CF{}, with 10\% increments.

\begin{figure}[!ht]
\centering
	\setlength{\fboxsep}{0pt}
	\setlength{\fboxrule}{0pt}%
\fbox{
\begin{subfigure}[t]{.47\linewidth}
	\includegraphics[scale=0.12]{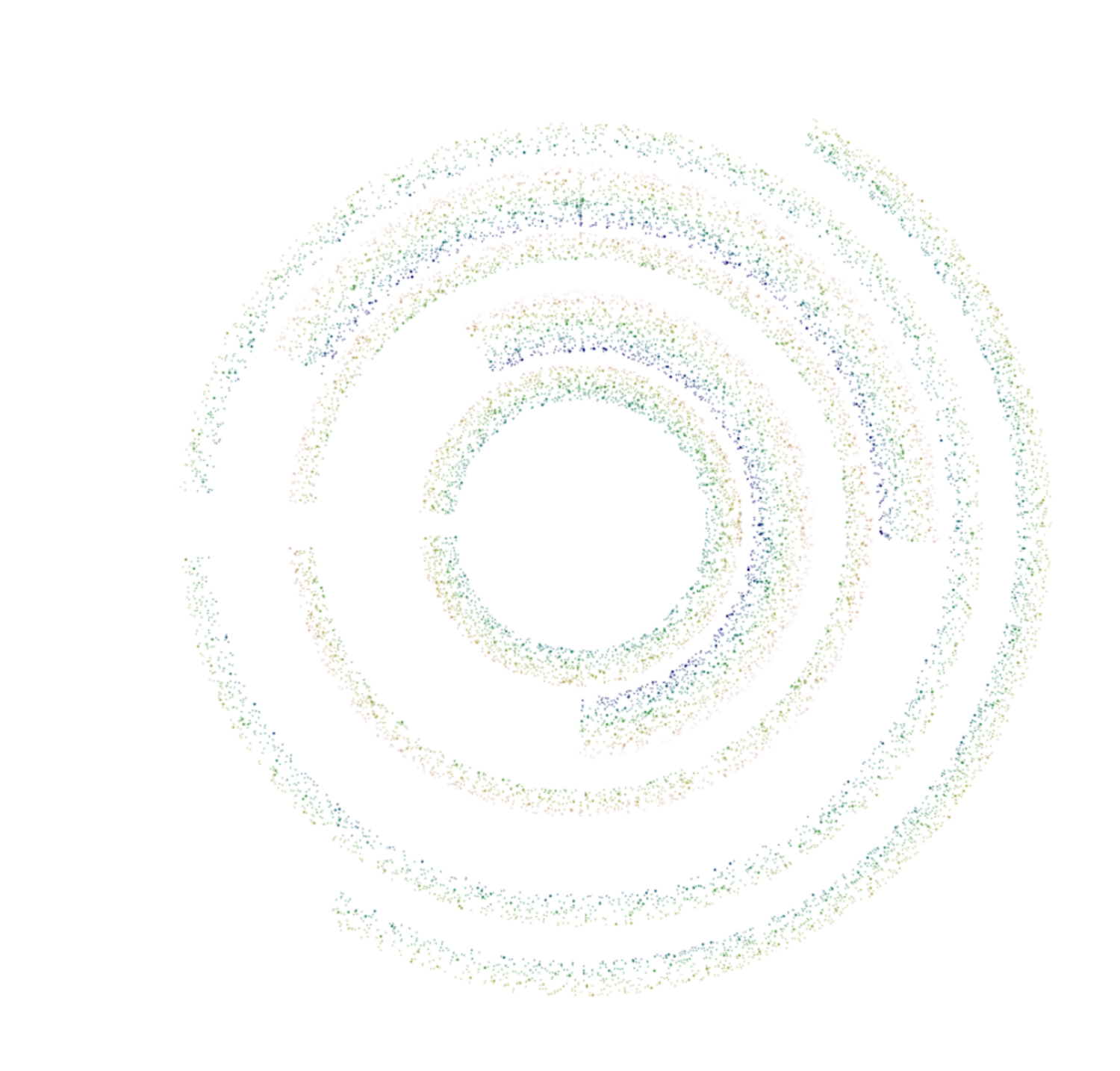}
	      \caption{Visualization of the set $A\setminus B$, here $A$ has a smaller uniform interval during discretization}
	      \label{fig:lessdisc}
\end{subfigure}
}
\fbox{
\begin{subfigure}[t]{.45\linewidth}
	 \includegraphics[scale=0.12]{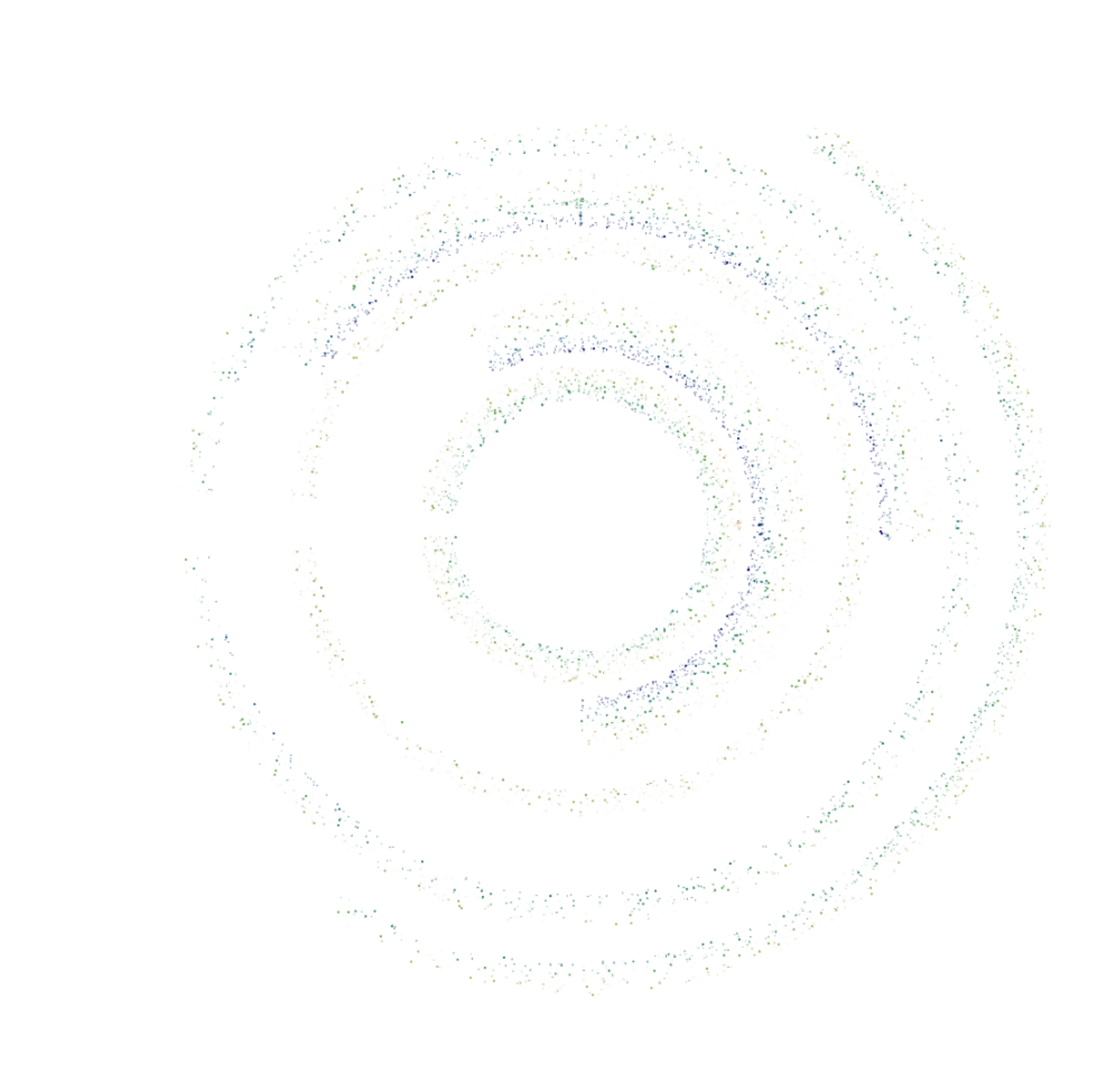}
	      \caption{Visualization of the set $A\setminus B$, here $A$ has a bigger uniform interval during discretization}
	      \label{fig:moredisc}
\end{subfigure}
}
\caption{The measure derived from $A\setminus B$ from the visualizations will change depending on the discretization applied to \CS{}, given that the number of pixels that are in $B^c$ and $A$ will change. In this case, $A$ is a representation of \figref{fig:wholeGuides4} that has $10\%$ of its states in collision, and $B$ is the original \figref{fig:wholeGuides4}}
\label{fig:discAccuracy}
	\vspace{-2mm}
\end{figure}

To assess the alignment between accuracy and our visualization approach, we conducted a set subtraction operation, represented as $A\setminus B=A\cap B^c$, between visualizations $A$ and $B$. During this operation, we replaced all pixels in the perturbed \CS{} visualizations that matched the non-perturbed \CS{} visualizations by setting their RGB values to white. Subsequently, we calculated the ratio of non-white pixels that remained after this replacement to the number of non-white pixels before the replacement in the perturbed visualization. This ratio served as our metric for measuring accuracy. We discretized the data in 500 different uniform intervals between $[-\pi,\pi]$. If the discretization is not made; we will under-estimate the accuracy when comparing two visualizations of the same \CS{} but with different sampled states. We can see an example of the discretization effect on $A\setminus B$ in \figref{fig:discAccuracy}.

	Given that each of the 100 different \CS{}s accuracies from visualizations are tested by measuring the correlation with the accuracy from the collision-checker, we present the average correlation using the Fisher-z transform; which results in $0.90\pm 0.099$ using the predicted standard error. This strong linear correlation \cite{10.5555/2502692} is expected, as we are directly plotting the sampled data projections.

Despite the difference in scale given by the discretization process, the strong correlation indicates that the proposed visualization is able to capture the change in accuracy when dealing with non-complete representations of \CF{}. This suggests that the visualization captures the accuracy of the collision checker in the context of high-dimensional \CS{}s and how the improvement of the \CFA{} is reflected in the visualization.

The proposed visualization is further tested to measure the difference between two \CS{}s, where one is a subset of the other, this measurement is not available using the original collision checker. In the context of image generation, there are various methods to compare images. The simplest one is to use the Mean Squared Error (MSE) per pixel. We can compare the visualization subsets of a \CS{} against the original visualization of \CS{} and measure how well the density of \CS{} is being reproduced. To evaluate this, we reduced the number of sampled points in each \CS{} from 90\% to 10\%, generating nine different subsets, and estimated the MSE between the original \CS{} with 100\% of its data points and its subsets.

\begin{table}[h!]
		\centering
		\caption{Mean and standard deviations of MSEs of the 100 different \CS{}s compared against their subsets with 90\% to 10\% of the original data samples (\% OD). The MSE derived from the visualization effectively gauges the extent to which the subset deviates from the complete model.}
\begin{tabular}{ c | c | c | c | c | c | c}
	\toprule
	\% OD
   & MSE & \% OD & MSE & \% OD& MSE\\
	\toprule
 90\% & 1.88 $\pm$ 0.01 &
 80\% & 1.92 $\pm$ 0.01 &
 70\% & 1.98 $\pm$ 0.01\\
 60\% & 2.05 $\pm$ 0.01 &
 50\% & 2.12 $\pm$ 0.01 &
 40\% & 2.21 $\pm$ 0.01\\
 30\% & 2.30 $\pm$ 0.01 &
 20\% & 2.40 $\pm$ 0.01 &
 10\% & 2.52 $\pm$ 0.01
\end{tabular}

	\label{table:mse}
\end{table}

	The visualization also encodes numerical information regarding how close the \CSA{} is to the original \CS{}'s collision checker, which can be analyzed using metrics from the field of computer vision, it is possible to measure the proximity of a representation model to the original \CS{}

In Table \ref{table:mse}, as the number of points forming the original \CS{} is decreased, the MSE increases. This implies that the visualization can effectively reflect how well a model is being represented based on the extent of \CF{} and it is able to encode numerical information regarding the accuracy of a \CSA{}, and even evaluate if only subsets of the orginal \CS{} are being represented. This is a capability not achievable when relying solely on a collision checker as a reference. Also, this result indicates that our visualization is able to capture coverage information.

	\section{Conclusions and Future Work}\label{sec:conclusions}
We have introduced a novel 2D visualization for high-dimensional \CS{}s of manipulator robots, aimed at facilitating the visual detection of regions where \CS{}s differ due to changes in their WS. Our findings demonstrate that our visualization approach effectively captures metrics like accuracy and error measurements between \CSA{}s, providing richer information compared to traditional approaches reliant solely on collision checking to assess differences between two \CS{}s, dimensional reduction or workspace projections; particularly in the context of \CS{} learning for SBMPs. This visualization offers valuable insights to machine learning practitioners in the field of robotics, enabling them to evaluate the quality of an approximation of a \CS{} by analyzing their visual differences in regions that are not restricted only to boundaries and shows how well the learned distribution covers the original one by comparing the amount of points that differ between the two distributions.

Future research will focus on adapting this visualization to various types of robots; such as mobile manipulators and conducting user studies. In such cases, it should be feasible to generate a visualization given that any bounded subinterval in $\Real$ is homeomorphic to $S^1$. However, further investigation is warranted for scenarios where movement might be unbounded or lacks a fixed reference for independent state visualization. The user interaction studies are very important to show the usability of the proposed visualization.

Additionally, a promising avenue for future exploration involves directly encoding the high-dimensional \CS{} into 2D visualizations for \CS{}-related tasks. Recent advancements in image-to-image generation algorithms, as demonstrated in the works of \cite{Wang2020NeuralRL} and \cite{JAS-2021-0110}, show promise in learning 2D representations by using visualizations of \CS{}, particularly when the dimension of \CS{} is also two. Leveraging such training approaches could extend the capabilities of these algorithms to high-dimensional \CS{} approximations, enabling the integration of state-of-the-art machine learning and computer vision techniques for \CS{} representation.

\end{document}